%% file: main.tex
\documentclass[runningheads]{llncs}

\usepackage{graphicx}
\usepackage{comment}
\usepackage{amsmath,amssymb}
\usepackage{color}
\usepackage{url}
\usepackage{hyperref}

\usepackage{subfigure}
\usepackage{soul,color}
\usepackage{amsmath,amssymb}
\usepackage{amsfonts}       
\usepackage{multirow}
\usepackage{makecell}
\usepackage[inline]{enumitem}
\usepackage{wrapfig}
\usepackage{pifont}
\usepackage{bm}
\usepackage[title]{appendix}

\newif\ifreview
\reviewfalse

\ifreview
	\usepackage{lineno}

	\linenumbers
\fi

\begin{document}


\def\SubNumber{33}

\def\GCPRTrack{Main Track}

\title{Multi-Attribute Open Set Recognition}

\ifreview
\else

    \author{Piyapat Saranrittichai\inst{1, 2}\orcidID{0000-0003-0620-7945} \and Chaithanya Kumar Mummadi \inst{1,2}\orcidID{0000-0002-1173-2720}\index{Mummadi, Chaithanya Kumar} \and Claudia Blaiotta \inst{1}\orcidID{0000-0003-2314-3939} \and Mauricio Munoz \inst{1}\orcidID{0000-0002-9520-4430} \and Volker Fischer\inst{1}\orcidID{0000-0001-5437-4030}}
	
	\authorrunning{P. Saranrittichai et al.}
	
	\institute{Bosch Center for Artificial Intelligence \and University of Freiburg}
\fi

\maketitle              

\input{sections/00_abstract}
\input{sections/01_introduction}
\input{sections/02_related_works}
\input{sections/03_methodology}
\input{sections/04_experiments}

\input{sections/05_conclusion}
\input{sections/06_acknowledgements}

%
%
%
%
\bibliographystyle{splncs04}
\bibliography{main.bbl}

\newpage
\appendix
\onecolumn
\begin{subappendices}
\renewcommand{\thesection}{\Alph{section}}
\input{sections/07_appendix}

\end{subappendices}

\end{document}

%% file: sections/00_abstract.tex
\begin{abstract}

Open Set Recognition (OSR) extends image classification to an open-world setting, by simultaneously classifying known classes and identifying unknown ones. While conventional OSR approaches can detect Out-of-Distribution (OOD) samples, they cannot provide explanations indicating which underlying visual attribute(s) (e.g., shape, color or background) cause a specific sample to be unknown. In this work, we introduce a novel problem setup that generalizes conventional OSR to a multi-attribute setting, where multiple visual attributes are simultaneously recognized. Here, OOD samples can be not only identified but also categorized by their unknown attribute(s). We propose simple extensions of common OSR baselines to handle this novel scenario. We show that these baselines are vulnerable to shortcuts when spurious correlations exist in the training dataset. This leads to poor OOD performance which, according to our experiments, is mainly due to unintended cross-attribute correlations of the predicted confidence scores. We provide an empirical evidence showing that this behavior is consistent across different baselines on both synthetic and real world datasets.

\keywords{Open Set Recognition  \and Multi-Task Learning \and Shortcut Learning.}
\end{abstract}

%% file: sections/01_introduction.tex
\section{Introduction}
\label{sec:introduction}

In recent years, deep learning techniques have been applied to a variety of complex classification tasks with great success. However, the majority of these methods rely on the closed-set assumption, where all classes in the test data are \emph{known} during training. As a result, such methods are poorly suited to more realistic scenarios where it is not possible to observe all classes during training. Open Set Recognition (OSR) \cite{walter2016towards} extends the standard classification task to handle \emph{unknown} classes that are not included in the training set - that is, an OSR model must discriminate known from unknown samples as well as classify known ones.

Previous research in the OSR domain has only considered a single classification task \cite{bendale2016towards,zhang2020hybrid,zhou2021learning,chen2021adversarial,guo2021conditional,vaze2021open}, where a network takes an image as an input and produces a label prediction, possibly accompanied by a confidence score (to make a decision on whether the input belongs to a known or unknown class). Predicting a single label, however, is not sufficient in some real world applications. For example, 
a robot picking up a red shirt needs to recognize multiple attributes of an object, such as its shape ("shirt") and its color ("red").

\input{figures/tex/figure_teaser}

The task of predicting multiple attributes simultaneously is closely related to Compositional Zero-Shot Learning (CZSL) \cite{misra2017red,nagarajan2018attributes,purushwalkam2019task,atzmon2020causal}, whose goal is to recognize unknown combinations of known attributes. However, one limitation of CZSL is that it assumes labels of input images to be already known. Therefore, when the model gets an input containing unknown attribute labels, it would incorrectly map them to known ones. In this work, we are interested in multi-attribute predictions, similar to CZSL, but we allow attributes of input images to be of unknown labels.


The scenario described above, can be cast as a multi-attribute OSR problem illustrated in Figure \ref{fig:teaser}. Given an image, the goal is to recognize multiple attributes (e.g., shape, color, background) at once. Additionally, if an attribute has an unknown value, the model should be able to identify it as such, irrespectively of whether the other attributes were seen or not before. We use the term \emph{attribute} to denote a generic visual property of image's which can be either low-level (i.e., shape, color, etc.) or high-level (i.e., object type, background, etc.) depending on the target tasks. Please note that our work is conducted under the assumption that the available attributes in a task are semantically independent (cannot describe the same aspects of an image). For example, in an object recognition task, texture and material are not semantically independent attributes, since the material attribute also contain texture information.



Several OSR approaches can naturally be extended to this novel setting. However, additional challenges arise in the case that certain combinations of known attributes are not available during training. This is especially likely when the number of feasible combinations is large. In case of missing combinations, spurious correlations among attributes are introduced leading to shortcut learning \cite{geirhos2020shortcut}. Our setting enables analyzing the effect of shortcut learning on OSR. In contrast to previous works in the shortcut learning literature, which analyze the effect of shortcuts only the recognition of known classes, our setting extends the analysis to unknown classes. We will also demonstrate that the multi-attribute extensions of common OSR techniques are shortcut vulnerable consistently across different approaches and datasets.


The main contributions of this work are as follows: (1)  We introduce the multi-attribute OSR setup, as a novel problem formulation promoting explainability. (2) We extend several OSR baselines to handle the multi-attribute setting. (3)  We demonstrate, with extensive empirical analyses, on both synthetic and realistic datasets, that many state-of-the-art OSR methods are vulnerable to shortcut learning. Specifically, we found OSR performance to be inversely correlated to the strength of dataset biases.

%% file: figures/tex/figure_teaser.tex
\begin{figure}[t]
    \centering
    \includegraphics[width=\textwidth]{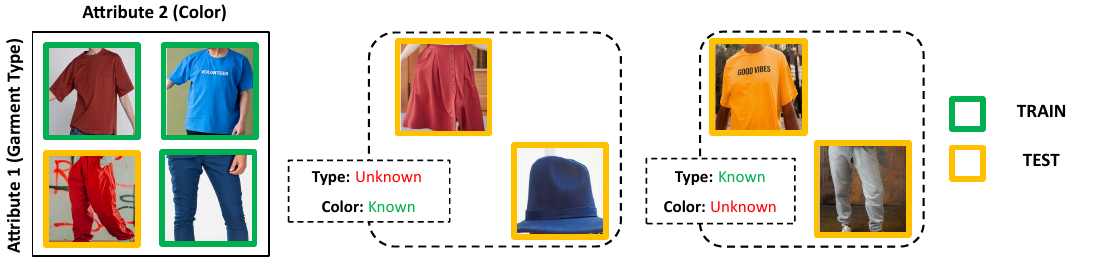}
    \caption{Illustration of our multi-attribute Open Set Recognition. Dataset contains images with attribute annotations (e.g., garment type and color). Samples with some known attribute combinations are available during training. During inference, networks need to be able to handle images containing both known and unknown attribute values. In case of input with unknown attribute values, the networks have to identify which attributes are actually of unknown values.}
    \label{fig:teaser}
\end{figure}

%% file: sections/02_related_works.tex
\section{Related Works}
\label{sec:related_works}

\subsection{Open Set Recognition}

Open-Set Recognition (OSR) is the problem of simultaneously classifying known samples and detecting unknown samples. OSR is conventionally a single-task problem, considering only one aspect of images (e.g., object type). With deep learning, the straightforward approach to solve OSR is Maximum Softmax Probability (MSP) applying softmax thresholding on the output of classification networks \cite{neal2018open}. \cite{bendale2016towards} shows that direct softmax thresholding does not minimize open space risk and proposes OpenMax as an alternative activation function that explicitly estimates the probability of a sample being unknown. \cite{shu2020p} models prototypes of individual object types to detect unknown categories. Other approaches instead rely on generative models to improve OSR \cite{guo2021conditional,zhou2021learning,yoshihashi2019classification,oza2019c2ae,sun2020conditional}. \cite{ge2017generative,neal2018open} generate synthetic unknown images to be used during training. \cite{zhang2020hybrid} directly estimates the likelihood of a sample being of a known class using a flow-based model. \cite{chen2020learning,chen2021adversarial} formulate constraints to maximize the difference between image features of known and unknown samples. Recently, \cite{vaze2021open} shows that network performance in OSR is heavily correlated with close-set accuracy and propose using Maximum Logit Score (MLS) for unknown class detection. These lines of work consider only object recognition but cannot recognize multiple attributes or infer which properties cause certain input images to be unknown. Some studies decompose feature space into multiple components based on image regions \cite{gillert2021towards} or attributes \cite{du2022class} to improve performance on fine-grained datasets. Still, their final task is that of single-task OSR. On the other hand, we study a generalized version of the OSR problem, in which multiple attributes are recognized simultaneously.

\subsection{Compositional Zero-Shot Learning}

Compositional Zero-Shot Learning (CZSL) is the the task of predicting both seen and unseen combinations of known attributes. VisProd is a na\"ive solution, which trains deep networks with cross-entropy losses to predict all attributes \cite{nagarajan2018attributes}. Recent works aim to model compatibility functions between attribute combinations and image features \cite{misra2017red,nagarajan2018attributes,purushwalkam2019task,atzmon2020causal,naeem2021learning,mancini2021open}. Our problem is closely related to CZSL as we also recognize multiple attributes at the same time. While CZSL assumes input images to have only known attribute values, we allow input images with unknown attribute values and would like to identify the unknown ones during inference.

\subsection{Shortcut Learning}

Shortcut Learning studies degradation of the recognition performance caused by spurious correlations \cite{geirhos2020shortcut,eulig2021diagvib}. \cite{geirhos2018imagenet} shows that real-world objects, in general, have shape-texture correlations and that deep networks trained with natural images tend to classify objects using the simpler visual clue, which is, in general, texture. \cite{hermann2020shapes} also shows that a trained model tends to extract information from visual features that are more linearly decodable. In contrast to most works on shortcut learning, which study its effects on the recognition of the known classes, we additionally study its effects on the recognition of the unknown classes.

%% file: sections/03_methodology.tex
\section{Problem Formulation}
\label{sec:problem_formulation}

In this work, we define \emph{attributes} as semantically independent visual properties of an image. For each attribute, we aim to predict its value or to recognize it as unknown if the value was never seen during training.

Our task formulation can be formally defined as follows. A sample from the training set contains an image $I^{i} \in \mathcal{I}^{train}$ as well as its attribute annotation $\bm{y}^{i} = (a_{1}^{i}, a_{2}^{i}, \ldots, a_{M}^{i}) \in \mathcal{Y}^{train} = \mathcal{A}^{k}_{1} \times \mathcal{A}^{k}_{2} \times \ldots \times \mathcal{A}^{k}_{M}$ where $M$ is the number of attributes and each $\mathcal{A}^{k}_{j}$ represents the set of all \emph{attribute values} of the $j$-th attribute seen during training. The superscript $k$ indicates that all attribute values available in the training set are of known values. On the other hand, a test sample can belong to either known or unknown values. In other words, $\mathcal{Y}^{test} = \left(\mathcal{A}^{k}_{1} \cup \mathcal{A}^{u}_{1}\right) \times \left(\mathcal{A}^{k}_{2} \cup \mathcal{A}^{u}_{2}\right) \times \ldots \times \left(\mathcal{A}^{k}_{M} \cup \mathcal{A}^{u}_{M}\right)$, where the superscript $u$ represents ones containing unknown values.

Given the training set, our goal is to estimate a function $f(I) = \left(\bm{\hat{y}}, \bm{\hat{s}}\right)$ that takes an image $I$ from a test sample as an input and predicts attribute values $\bm{\hat{y}} \in \mathcal{Y}^{train}$ as well as attribute-wise confidence scores $\bm{\hat{s}} = \left(\hat{s}_1, \hat{s}_2, \ldots, \hat{s}_M \right)$. These confidence scores will be used to distinguish between known and unknown attribute values by thresholding. In our work, higher confidence score indicates higher likelihood for a sample to be of a known attribute value.

During training, some attribute combinations can be missing (see Figure \ref{fig:correlation_configurations}). On the other hand, any attribute combinations can be seen during testing, including the combinations containing unknown attribute values. We define the term \emph{Open-Set OOD for $\mathcal{A}_{m}$} ($OOD^O_m$) to represent a sample whose attribute values are partially unknown on the $m$-th attribute. Additional terms indicating samples with different categories of attribute combinations are defined in Figure \ref{fig:sample_definitions}. 

\input{figures/tex/figure_sample_definitions}

\section{Extension from common OSR approaches}
\label{sec:method_extension}

\input{figures/tex/figure_architectures}

For conventional single-task OSR, which is a special case of our setting where $M = 1$, model solves a single task and predicts one class label and one confidence score. These predictions are typically computed by a network combining a feature extractor (to compute intermediate representations) with a prediction head (to compute final predictions) as shown in Figure \ref{fig:architectures_conventional}. Given a feature extractor $g$ and a prediction head $h$, the function $f$ can be written as $f(I) = h(g(x))$, where $g(I) = z$ is the intermediate feature representation.


In this work, we study a simple generalization of single-task OSR methods, which is directly applicable to several state-of-the-art approaches.
To extend a single-task model, we opt to use only a single feature extractor $g$ similar to the case of conventional OSR. However, $h$ is extended to have multiple heads $(h_1, h_2, \ldots, h_M)$ to support simultaneous prediction of multiple attributes. This extension can be seen in Figure \ref{fig:architectures_multi}.








%% file: figures/tex/figure_sample_definitions.tex
\begin{figure}[t]
    \centering
    \includegraphics[width=0.5\textwidth]{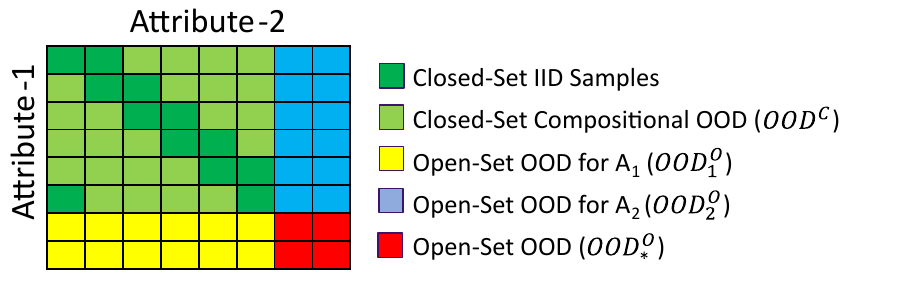}
    \caption{Definitions of samples which can be found during inference. Samples with (dark or light) green colors represent samples whose attribute values are all \emph{known}. These values can be either the seen combinations during training (dark green) or the unseen ones (light green). Yellow, blue and red boxes represent attribute combinations containing \emph{unknown} attribute values. The yellow and blue boxes are partially unknown as only the values of attribute 1 ($OOD^O_1$) or attribute 2 ($OOD^O_2$) are unknown respectively.}
    \label{fig:sample_definitions}
\end{figure}

%% file: figures/tex/figure_architectures.tex
\begin{figure} [t]
    \centering
    \subfigure[Conventional OSR]{
        \includegraphics[width=0.4\textwidth]{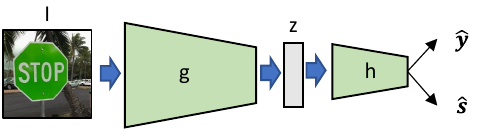}
        \label{fig:architectures_conventional}
    }
    \subfigure[Multi-attribute OSR ($M = 2$)]{
        \includegraphics[width=0.4\textwidth]{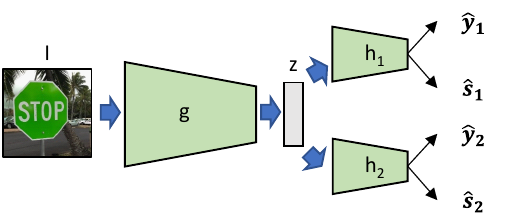}
        \label{fig:architectures_multi}
    }
    \caption{Illustration of generic architectures for single- and multi-attribute OSR.}
    \label{fig:architectures}
\end{figure}


%% file: sections/04_experiments.tex
\section{Experiments}
\label{sec:experiments}

In this section, we present a set of experiments and analysis to understand the behavior of different methods on multi-attribute OSR setting. We begin by describing our experimental setup and then present quantitative evaluation of various approaches on synthetic datasets. Additionally, we uncover a possible mechanism by which shortcuts can affect OSR performance, which relates to undesired correlation of confidence scores across attributes. Lastly, we show that our findings can also be observed in a more realistic dataset.

\subsection{Experiment Setup}
\label{sec:experiments_setup}

In this subsection, details of our experiment setup will be described. In particular we will present the datasets, baselines and metrics used in our study.

\subsubsection{Datasets}

\input{figures/tex/figure_dataset_visualization}

Datasets with multiple attribute annotations are required for multi-attribute OSR setting. Additionally, to enable shortcut analysis, we choose datasets where the available attributes have different complexity (i.e., in term of linear decodability as in \cite{hermann2020shapes}). In the presence of shortcuts, we expect neural networks to rely heavily on simpler but less robust visual clues. In this work, we consider datasets where two attributes are annotated ($M = 2$). The first attribute $\mathcal{A}_{c}$ is intended to be complex (e.g., shape, object type) while the second attribute $\mathcal{A}_{s}$ is intended to be simple (e.g., color, background). The datasets used in this work are as follows (example images are shown in Figure \ref{fig:dataset_visualization}):

\begin{itemize}
    \item Color-MNIST: This dataset contains MNIST digits with different colors. We construct this dataset based on the DiagVib framework \cite{eulig2021diagvib}. Attributes available in this dataset are digit shape and color.
    \item Color-Object: This dataset is inspired by \cite{ahmed2020systematic} to simulate more realistic settings while full-control over dataset biases is still maintained. Each image sample contains an object from COCO \cite{lin2014microsoft} over a background of single color. Attributes available in this dataset are object type and background color respectively.
    \item Scene-Object: Similar to Color-Object, each image sample still contains a COCO object. However, instead of using color as background, scene images from \cite{zhou2017places} are used.
    \item UT-Zappos: This dataset contains images of shoes annotated with two attributes (shoe material and type) \cite{yu2014fine,yu2017semantic} which is commonly used as a CZSL benchmark.
\end{itemize}

We split datasets mentioned above so that some attribute values are kept unknown during training for OSR evaluation. Details of each dataset split will be presented in appendix \ref{appendix:dataset_details}. 

\subsubsection{Baselines}

All baselines considered in this work comprise a feature extraction backbone ($g$) and multiple prediction heads $(h_1, h_2, \ldots, h_M)$, as introduced in section \ref{sec:method_extension}. The main difference among baselines lies in the modeling of loss functions and retrieval of confidence scores. Baselines considered in this work are:

\begin{itemize}
    \item Maximum Softmax Probability (MSP) \cite{neal2018open}: This standard baseline trains networks by minimizing a sum of cross-entropy losses across all prediction heads. The confidence score for each attribute is simply the maximum softmax value from its prediction head.
    \item OpenMax \cite{bendale2016towards}: The training procedure of this baseline is similar to MSP. However, the final confidence score of each attribute will be calibrated to reduce probability of the predicted labels based on the distance between the estimated logit vector and its corresponding mean vector.
    \item ARPL \cite{chen2021adversarial}: This baseline trains the model so that feature representations of potential unknown samples are far away from the ones of known samples. The suffix +CS is appended when \emph{confusing samples} are also generated and used for training.
    \item Maximum Logit Score (MLS) \cite{vaze2021open}: This baseline uses maximum logit value before softmax activation from each prediction head as the confidence score. This prevents certain information being discarded by softmax normalization.
\end{itemize}

Model architectures and hyperparameters used for each dataset will be presented in appendix \ref{appendix:implementation_details}.

\subsubsection{Metrics}

For the evaluation, we extend conventional OSR metrics for multi-task setting. In contrast to conventional OSR, we have multiple prediction heads so that the evaluation will be performed per head. We consider the average metrics over all heads to represent overall performance. Two main metrics adopted here are Open-Set Classification Rate (OSCR) and Area Under the Receiver Operating Characteristic (AUROC) similar to \cite{chen2021adversarial,vaze2021open}. Quantitative results reported in this section are averaged across three different random seeds.

\subsection{Shortcut Analysis in Multi-attribute OSR}
\label{subsec:experiments_synthetic}

\input{figures/tex/figure_correlation_configurations}

In this section, we study effects of shortcut learning on synthetic datasets in which we can control the attribute combinations seen during training. A low number of seen combinations can amplify the effects of shortcut learning. As illustrated in Figure \ref{fig:correlation_configurations}, we utilize three dataset configurations: (1) \textbf{Correlated (C)}: With this configuration, the number of seen combination is minimal. In other words, there is a one-to-one mapping between each available attribute. (2) \textbf{Semi-Correlated (SC)}: With this configuration, additional combinations are introduced so that a trained network has enough information to distinguish two visual attributes. The number of combinations in this configuration is two times higher than in the Correlated configuration. (3) \textbf{Uncorrelated (UC)}: All combinations are available.

We control dataset generation such that the number of samples are balanced across all attribute combinations. During testing, for each attribute, the number of the unknown attribute values is the same as the number of the known attribute values. In addition, the test combinations are also balanced so that the random chance of detecting an unknown attribute is 50\% per attribute. More details of the splits are presented in appendix \ref{appendix:dataset_details}.

\input{tables/table_correlations_oscr_avg}

\input{figures/tex/figure_oscr_graph}

Table \ref{table:correlations_oscr_avg} shows OSCR across different approaches and dataset configurations (i.e., Uncorrelated (UC), semi-correlated (SC) and correlated (C)). We notice that, in uncorrelated cases, OSCR values are consistently the highest across all baselines and datasets. These values correspond to the ideal case that there is no dataset bias caused by spurious attribute correlations (low shortcut opportunities). In contrast, OSR performance degrades as soon as correlations are introduced, such as in the SC and C configurations. This suggests that attribute correlations in the SC and C cases can induce shortcut learning during training. Similar trends can also be observed with AUROC metrics (see appendix \ref{appendix:quantitative_auroc_avg}).

In order to investigate models's behavior in the presence of shortcut opportunities, we look at attribute-wise OSCR for the complex attribute ($\mathcal{A}_c$) and the simple attribute ($\mathcal{A}_s$). Figure \ref{fig:oscr_graph} presents OSCR of $\mathcal{A}_c$ and $\mathcal{A}_s$ from different baselines on Color-MNIST dataset. According to the results in the Figure, spurious correlations do not have the same effects to both attributes. Specifically, the reduction of OSCR mainly occurs on $\mathcal{A}_c$ but not on $\mathcal{A}_s$ (For some baselines, OSCR values of $\mathcal{A}_s$ do not even reduce). A possible explanation of this behavior is that, during correlated training, a network tends to rely more heavily on the simple attribute compared to the complex one when both attributes are predictive of the model's target. Similar insights were reported by previous works on shortcut learning \cite{geirhos2020shortcut,hermann2020shapes}. More evaluation details and discussions regarding to the attribute-wise evaluations will be presented in the appendix \ref{appendix:quantitative_attribute_wise_performance}.


In addition to the baseline extension method described in section \ref{sec:method_extension}, we also study another simple approach that simply duplicates the whole model for each attribute (represented with a -D appended after the approach names) resulting in models with two times many parameters (in the case of two attributes). According to the results in Table \ref{table:correlations_oscr_avg}, the general trend of the performance degradation due to spurious correlations is still hold. Throughout this work, we opt to perform our analysis mainly based on the extension introduced in section \ref{sec:method_extension}.



\subsection{Effects of Shortcut on Confidence Scores}
\label{subsec:experiments_confidence_scores}

The results presented in the previous section indicate that spurious correlations can degrade multi-attribute OSR performance, as measured using common OSR metrics. Even though these metrics are useful for comparing models, they do not uncover the underlying mechanism by which the detection of unknown samples degrades. Since confidence scores are the main output used to distinguish between known and unknown samples, we will explore how shortcut learning affects the confidence scores in this section.

\input{figures/tex/figure_ideal_confidence_scores}

We will visualize the behavior of confidence scores as a heatmap $\mathcal{M}$ similar to Figure \ref{fig:ideal_confidence_scores}. A value $\mathcal{M}_{ij}$ in the $(i, j)$ cell is the average confidence score from the prediction head of the $i$-th attribute given input sample of type $j$ (1=Known, 2=Open-Set OOD for attribute 1 ($\mathcal{A}_c$), 3=Open-Set OOD for attribute 2 ($\mathcal{A}_s$)). The ideal behaviour is that, if an input sample has a known attribute value for the attribute $i$, the confidence score of the $i$-th head should be similar to the case of known inputs (i.e., $\mathcal{M}_{11} \approx \mathcal{M}_{13}$ and $\mathcal{M}_{21} \approx \mathcal{M}_{22}$). On the other hand, if an input sample is unknown with respect to the $i$-attribute, the $i$-th confidence score should be lower (i.e., $\mathcal{M}_{11} > \mathcal{M}_{12}$ and $\mathcal{M}_{21} > \mathcal{M}_{23}$), in which case by thresholding the score, one can identify the unknown attribute value. Ideally, the confidence score of one attribute should not be impacted by the attribute value of another attribute. We called this property of the confidence score as \emph{cross-attribute independence}.

\input{figures/tex/figure_cconf_dv}

\input{figures/tex/figure_cconf_obc_color}

We test the standard MSP baseline on uncorrelated, semi-correlated and correlated setting. The result of the Color-MNIST and Color-Object dataset shown in Figure \ref{fig:cconf_dv} and Figure \ref{fig:cconf_obc_color}. Firstly, we notice that, in the uncorrelated case, confidence scores are more or less similar to the ideal behaviour mentioned above (e.g., see Figure \ref{fig:cconf_dv_softmax_uc} and \ref{fig:cconf_obc_color_softmax_uc}). However, as the correlation is introduced, a cross-attribute dependency of the confidence scores can be observed. For example, considering the semi-correlated case in Figure \ref{fig:cconf_obc_color_softmax_sc}, confidence score of the first attribute (object type) drops when the second attribute (background) is unknown ($\mathcal{M}_{11} > \mathcal{M}_{13}$). This indicates unintended dependency between confidence score of the first attribute and the attribute value of the second attribute. In other words, the network incorrectly uses the information of second attribute to determine if the first attribute is unknown or not. This demonstrates one scenario that shortcut can affect the recognition performance of OOD. In addition, considering edge cases in which all attributes are correlated in Figure \ref{fig:cconf_dv_softmax_c} and \ref{fig:cconf_obc_color_softmax_c}, average confidence scores of the first attribute are unchanged regardless of whether actual attribute values are unknown or not (i.e., $\mathcal{M}_{11} = \mathcal{M}_{12}$). 
The network only use the information of the second attribute to detect the unknown of both attributes. This is an obvious sign of shortcuts similar to the semi-correlated case mentioned before.



\subsection{Measuring Explainability for Unknown Samples}

\input{figures/tex/figure_oe_dv}

\input{figures/tex/figure_oe_obc_color}

As mentioned earlier, one benefit of our task is that, in contrast to conventional OSR or anomaly detection, models will be able to provide meaningful attribute-level explanations of why certain samples are identified as unknown. In other words, because models categorize each attribute separately as known or unknown, they can also explain which underlying attribute(s) are responsible for an input sample likely being unknown.

In this regard, we evaluate explainability using a confusion matrix $\mathcal{C}$, similarly to closed-set evaluation. However, instead of grouping samples by class labels, we group samples based on whether they are known or unknown with respect to specific attributes. Each row and column index correspond to a type of sample, either known, Open-Set OOD for attribute 1, Open-Set OOD for attribute 2 or Open-Set OOD (refer to the definitions in Figure \ref{fig:sample_definitions}). The horizontal and vertical axes correspond to prediction and ground truth respectively. A value in each cell indicates the percentage of samples from the ground truth group corresponded to the row index which the network predicts to the group corresponded to the column index. The sum of values in each row is normalized to 100. Ideally, we would like to get high values along the diagonal and low values on off-diagonal cells which indicates the prediction of the correct groups. It should be noted that, unlike previous experiments, the evaluation in this section needs thresholding to identify the unknowns. Here, we pick thresholds that maximize micro-F-Measure metric \cite{geng2020recent}.

Results on the Color-MNIST and Color-Object datasets are shown in Figure \ref{fig:oe_dv} and \ref{fig:oe_obc_color}. In the uncorrelated case, we obtain high values along the diagonal, thus indicating good explainability of OOD samples (see Figure \ref{fig:oe_dv_softmax_uc} and \ref{fig:oe_obc_color_softmax_uc}). Shortcuts do, as to be expected, degrade model's ability to explain an OOD outcome, as shown by the results in the semi-correlated and correlated cases. In Figure \ref{fig:oe_obc_color_softmax_sc}, the high value in the off-diagonal cell $\mathcal{C}_{12}$ indicates poor performance to recognize unknown for the complex attribute (i.e., object type in this case). This result is consistent as in section \ref{subsec:experiments_confidence_scores} since shortcuts make confidence scores of complex attributes less distinguishable for its own attribute. Recognition performance of the complex attribute is also poor even for the known attribute values, therefore, the optimal threshold get less reward from identifying a sample as the known. Thresholding is then likely to output more unknown predictions. This behaviour can be more clearly seen in the fully-correlated case (Figure \ref{fig:oe_dv_softmax_c} and \ref{fig:oe_obc_color_softmax_c}), where complex attributes for all samples are identified as unknown. From all of our experiments, we can conclude that shortcuts affect performance to recognize OOD and explanability of the models.

\subsection{Evaluation on a Realistic Dataset}
\label{subsec:experiments_real}

\input{figures/tex/figure_evaluation_ut_zappos}

In previous sections, we conducted experiments on well-controlled synthetic datasets and empirically shows effects of shortcut learning on multi-attribute OSR. In this section, we will demonstrate that our findings on synthetic datasets can also be observed in realistic settings. In this regard, we use an open set split based on UT-Zappos dataset \cite{yu2014fine,yu2017semantic}, a common dataset for CZSL. The dataset contains shoe images annotated with shoe materials and types. In the context of UT-Zappos, shoe material is relatively fine-grained compared to shoe type (as seen in Figure \ref{fig:dataset_visualization_ut_zappos}). Therefore, we consider shoe material as the complex attribute ($\mathcal{A}_c$) and shoe type as the simple attribute ($\mathcal{A}_s$).

Figure \ref{fig:evaluation_ut_zappos_auroc_avg} shows AUROC on UT-Zappos from various baselines. The general trend follows synthetic scenarios that AUROC values tend to be lower for $\mathcal{A}_c$ and higher for $\mathcal{A}_s$. We also look at the cross-attribute confidence scores for MSP as in Figure \ref{fig:evaluation_ut_zappos_cconf_msp}. According to the Figure, the scores are similar for the first and second column ($\mathcal{M}_{11} \approx \mathcal{M}_{12}$ and $\mathcal{M}_{21} \approx \mathcal{M}_{22}$) where shoe types are known, while, the scores drop on the third column ($\mathcal{M}_{13}$ and $\mathcal{M}_{23}$) where shoe types are unknown. This indicates that shoe types are mainly determined the output of unknown detection even for the shoe material prediction head. In other words, models take $\mathcal{A}_s$ as shortcuts for unknown detection of $\mathcal{A}_c$ demonstrating the same observation as in the semi-correlated cases of section \ref{subsec:experiments_confidence_scores}. This shows that our findings are also generalized to realistic scenarios.

%% file: figures/tex/figure_dataset_visualization.tex
\begin{figure*} [t]
    \centering
    \subfigure[Color-MNIST]{
        \includegraphics[width=0.22\textwidth]{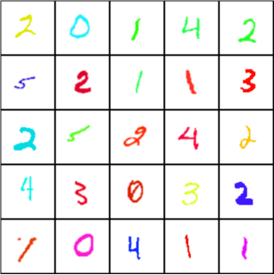}
        \label{fig:dataset_visualization_color_mnist}
    }
    \subfigure[Color-Object]{
        \includegraphics[width=0.22\textwidth]{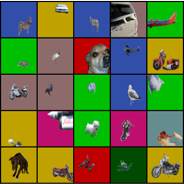}
        \label{fig:dataset_visualization_color_objects}
    }
    \subfigure[Scene-Object]{
        \includegraphics[width=0.22\textwidth]{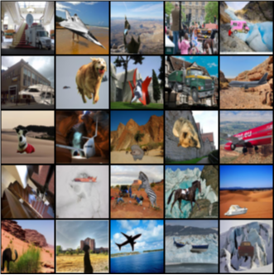}
        \label{fig:dataset_visualization_scene_objects}
    }
    \subfigure[UT-Zappos]{
        \includegraphics[width=0.22\textwidth]{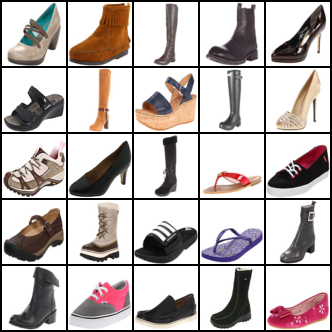}
        \label{fig:dataset_visualization_ut_zappos}
    }

    \caption{Images of example samples from datasets used in this work.}
    \label{fig:dataset_visualization}
\end{figure*}

%% file: figures/tex/figure_correlation_configurations.tex
\begin{figure}[t]
    \centering
    \includegraphics[width=0.8\textwidth]{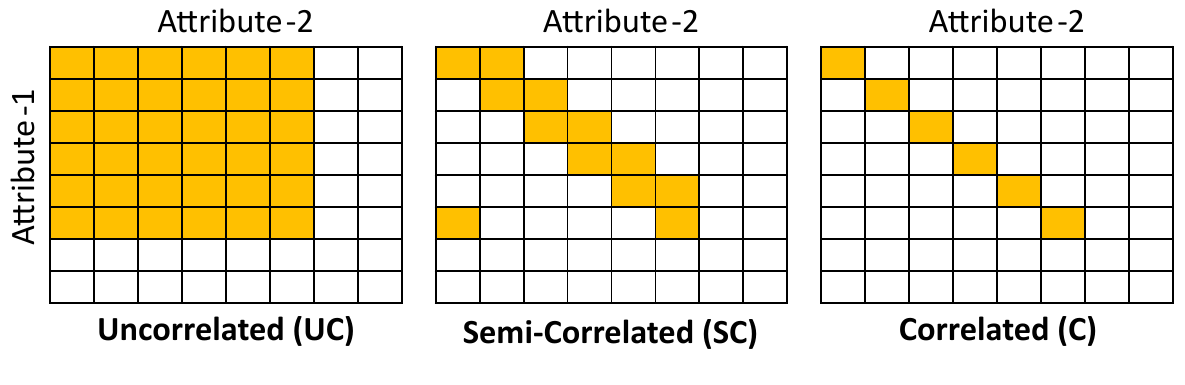}
    \caption{Training dataset configurations. Each box represents an attribute combination whose yellow color indicates its presence during training. These three configurations provide different degrees of spurious correlations to simulate different shortcut opportunities.}
    \label{fig:correlation_configurations}
\end{figure}

%% file: tables/table_correlations_oscr_avg.tex
\begin{table*}[t]
\caption{Average OSCR (from all attributes) on Color-MNIST, Color-Object and Scene-Object.}
\centering
\fontsize{9}{11}\selectfont
\begin{tabular}{l|ccc|ccc|ccc}
\Xhline{4\arrayrulewidth}
\multirow{2}{*}{Approach} & \multicolumn{3}{c|}{Color-MNIST} & \multicolumn{3}{c|}{Color-Object} & \multicolumn{3}{c}{Scene-Object} \\ 
& UC & SC & C & UC & SC & C & UC & SC & C \\ \hline
 MSP       & $80.5$       & $56.0$                 & $42.9$     & $76.2$                     & $58.4$                               & $53.0$                   & $39.7$                     & $32.1$                               & $29.3$                   \\
 OpenMax   & $79.1$       & $52.9$                 & $45.2$     & $70.2$                     & $45.8$                               & $25.0$                   & $35.5$                     & $18.4$                               & $13.1$                   \\ 
 ARPL      & $80.8$       & $54.0$                 & $49.2$     & $77.3$                     & $54.4$                               & $36.0$                   & $42.5$                     & $34.5$                               & $29.9$                   \\ 
 ARPL+CS   & $78.7$       & $52.7$                 & $49.9$     & $77.6$                     & $53.9$                               & $38.5$                   & $44.7$                     & $34.8$                               & $28.3$                   \\
 MLS       & $76.5$       & $44.5$                 & $36.0$     & $74.9$                     & $57.0$                               & $51.6$                   & $40.4$                     & $32.2$                               & $29.4$                   \\ \hline
 MSP-D     & $80.0$       & $58.3$                 & $46.4$     & $75.0$                     & $61.7$                               & $52.4$                   & $42.9$                     & $35.5$                               & $28.8$                   \\
 OpenMax-D & $76.8$       & $49.1$                 & $43.8$     & $75.2$                     & $48.5$                               & $22.6$                   & $38.6$                     & $16.6$                               & $12.2$                   \\ 
 ARPL-D    & $83.6$       & $60.3$                 & $45.8$     & $79.1$                     & $59.2$                               & $49.8$                   & $44.0$                     & $33.1$                               & $26.3$                   \\ 
 \Xhline{4\arrayrulewidth}
\end{tabular}
\label{table:correlations_oscr_avg}
\end{table*}

%% file: figures/tex/figure_oscr_graph.tex
\begin{figure} [t]
    \centering
    \subfigure[Complex Attribute]{
        \includegraphics[width=0.45\textwidth]{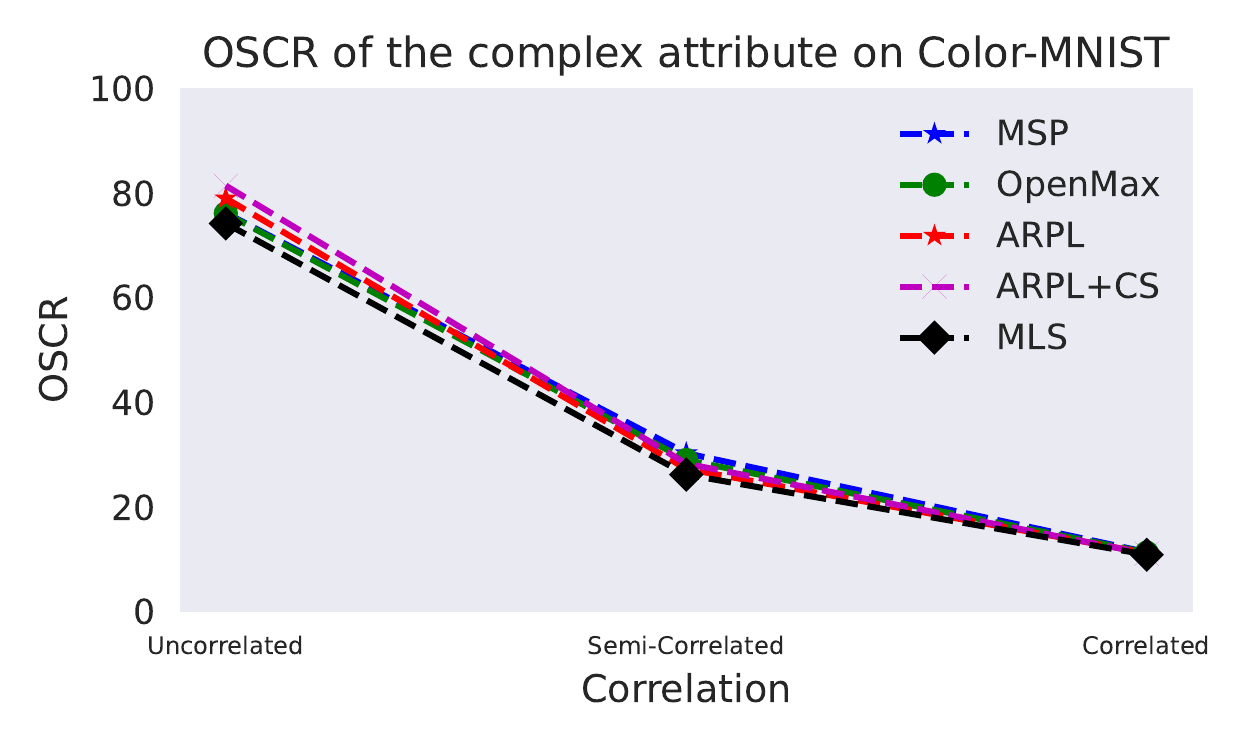}
        \label{fig:oscr_graph_complex}
    }
    \subfigure[Simple Attribute]{
        \includegraphics[width=0.45\textwidth]{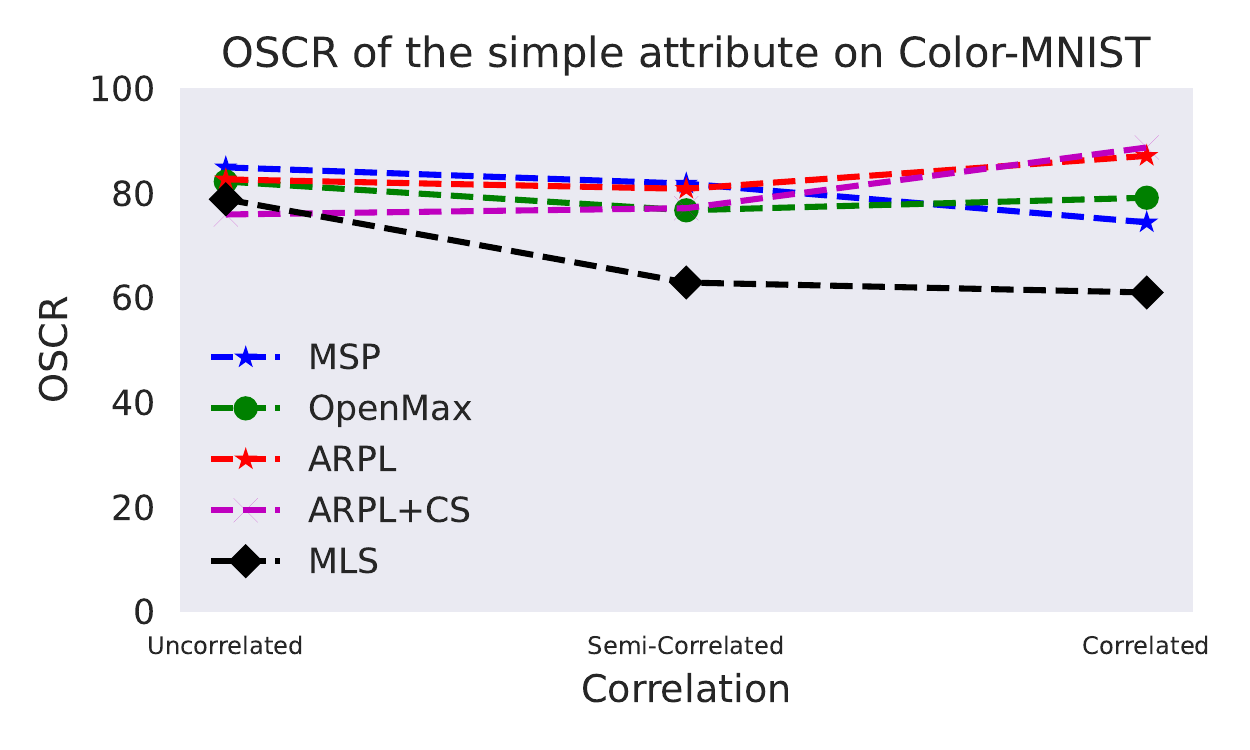}
        \label{fig:oscr_graph_simple}
    }
    \caption{Attribute-wise OSCR evaluation on Color-MNIST with different dataset configurations.}
    \label{fig:oscr_graph}
\end{figure}

%% file: figures/tex/figure_ideal_confidence_scores.tex
\begin{figure}[t]
    \centering
    \includegraphics[width=0.8\textwidth]{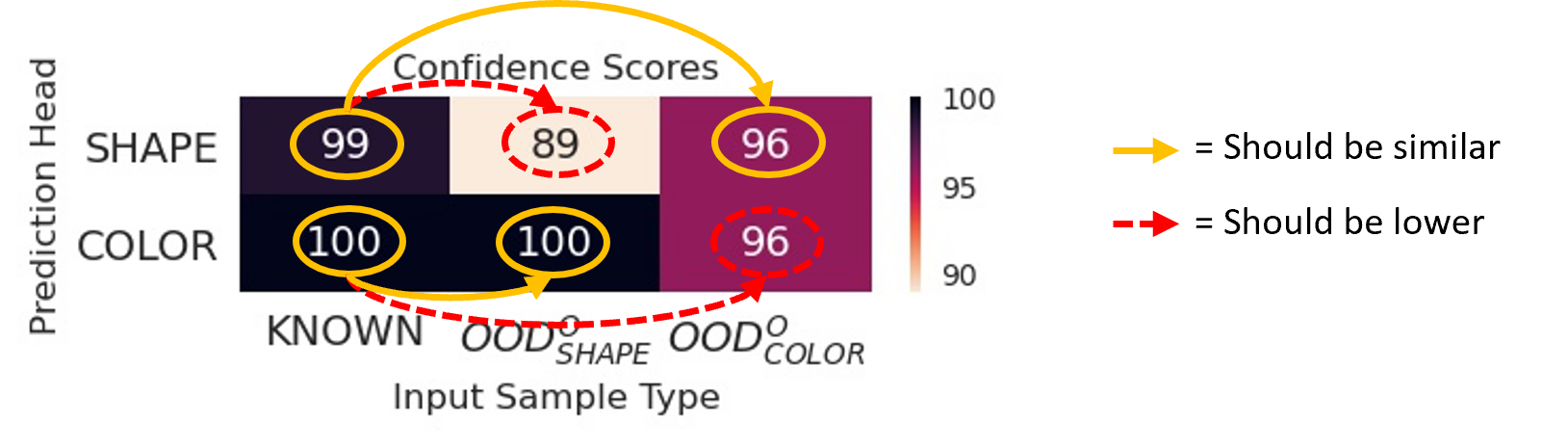}
    \caption{Visualization of cross-attribute confidence scores}
    \label{fig:ideal_confidence_scores}
\end{figure}

%% file: figures/tex/figure_cconf_dv.tex
\begin{figure*} [t]
    \centering
    \subfigure[Uncorrelated (UC)]{
        \includegraphics[width=0.30\textwidth]{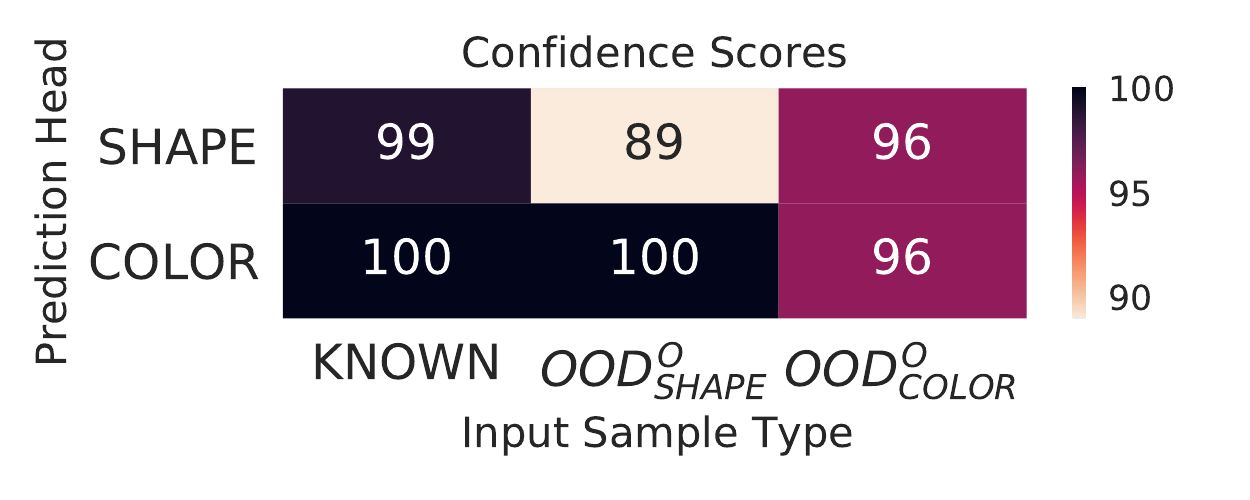}
        \label{fig:cconf_dv_softmax_uc}
    }
    \subfigure[Semi-Correlated (SC)]{
        \includegraphics[width=0.30\textwidth]{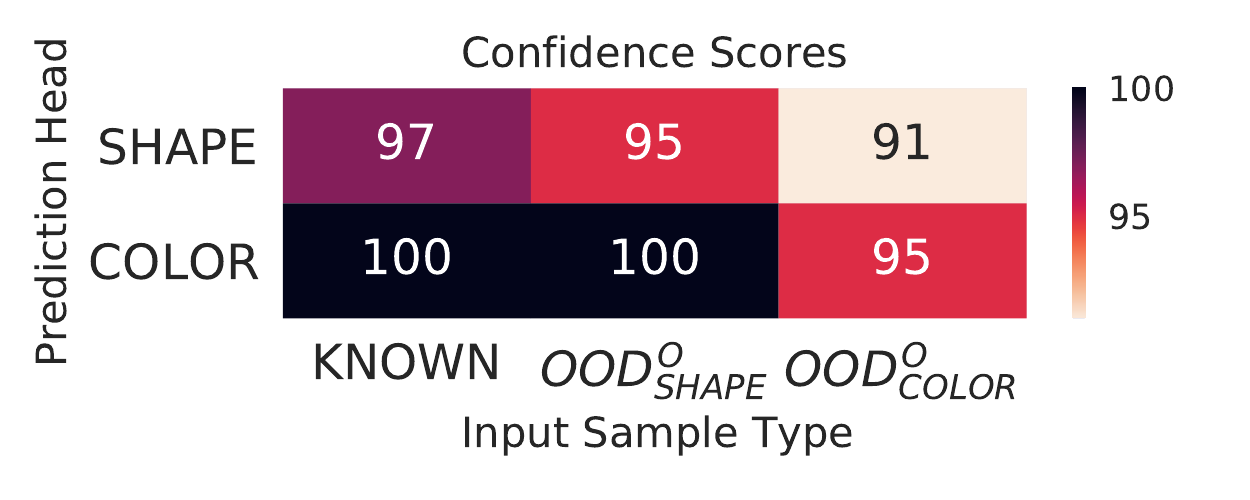}
        \label{fig:cconf_dv_softmax_sc}
    }    
    \subfigure[Correlated (C)]{
        \includegraphics[width=0.30\textwidth]{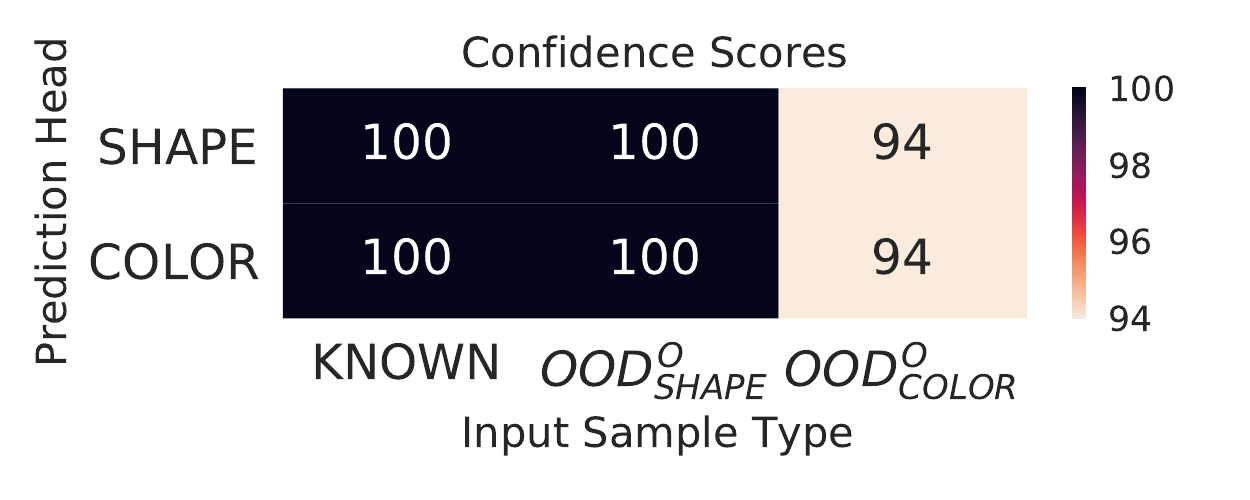}
        \label{fig:cconf_dv_softmax_c}
    }
    
    \caption{Cross-attribute confidence scores of MSP on different configurations of the Color-MNIST dataset.}
    \label{fig:cconf_dv}
\end{figure*}

%% file: figures/tex/figure_cconf_obc_color.tex
\begin{figure*} [t]
    \centering
    \subfigure[Uncorrelated (UC)]{
        \includegraphics[width=0.3\textwidth]{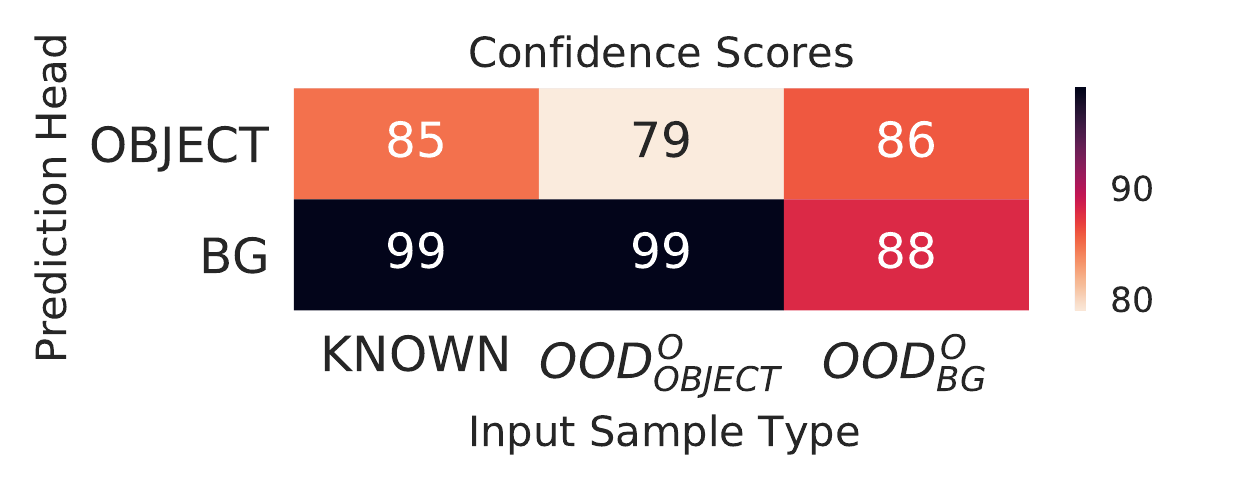}
        \label{fig:cconf_obc_color_softmax_uc}
    }
    \subfigure[Semi-Correlated (SC)]{
        \includegraphics[width=0.3\textwidth]{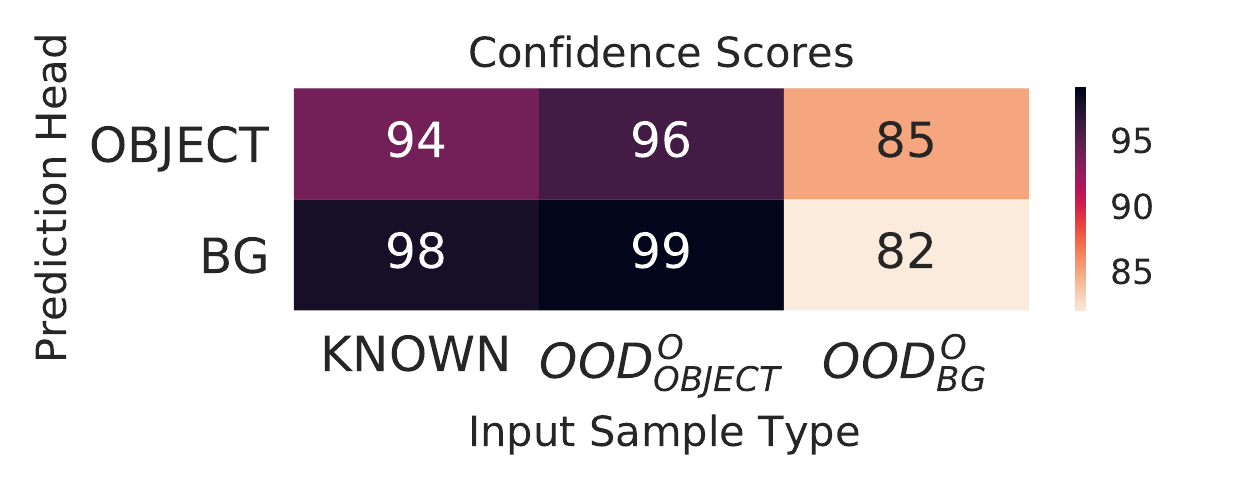}
        \label{fig:cconf_obc_color_softmax_sc}
    }    
    \subfigure[Correlated (C)]{
        \includegraphics[width=0.3\textwidth]{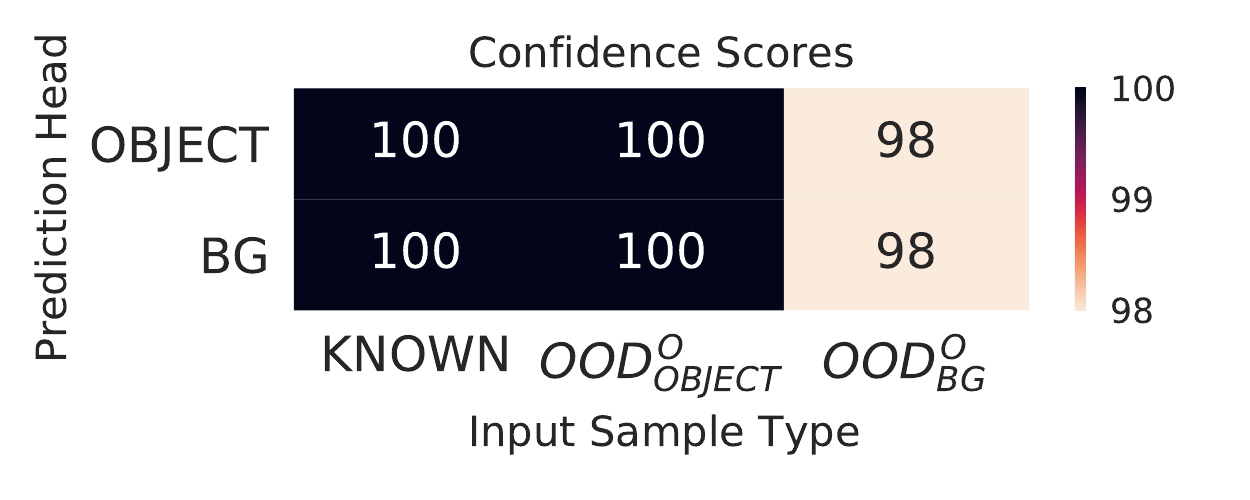}
        \label{fig:cconf_obc_color_softmax_c}
    }

    \caption{Cross-attribute confidence scores of MSP on different configurations of Color-Object dataset.}
    \label{fig:cconf_obc_color}
\end{figure*}

%% file: figures/tex/figure_oe_dv.tex
\begin{figure*} [ht]
    \centering
    \subfigure[Uncorrelated (UC)]{
        \includegraphics[width=0.3\textwidth]{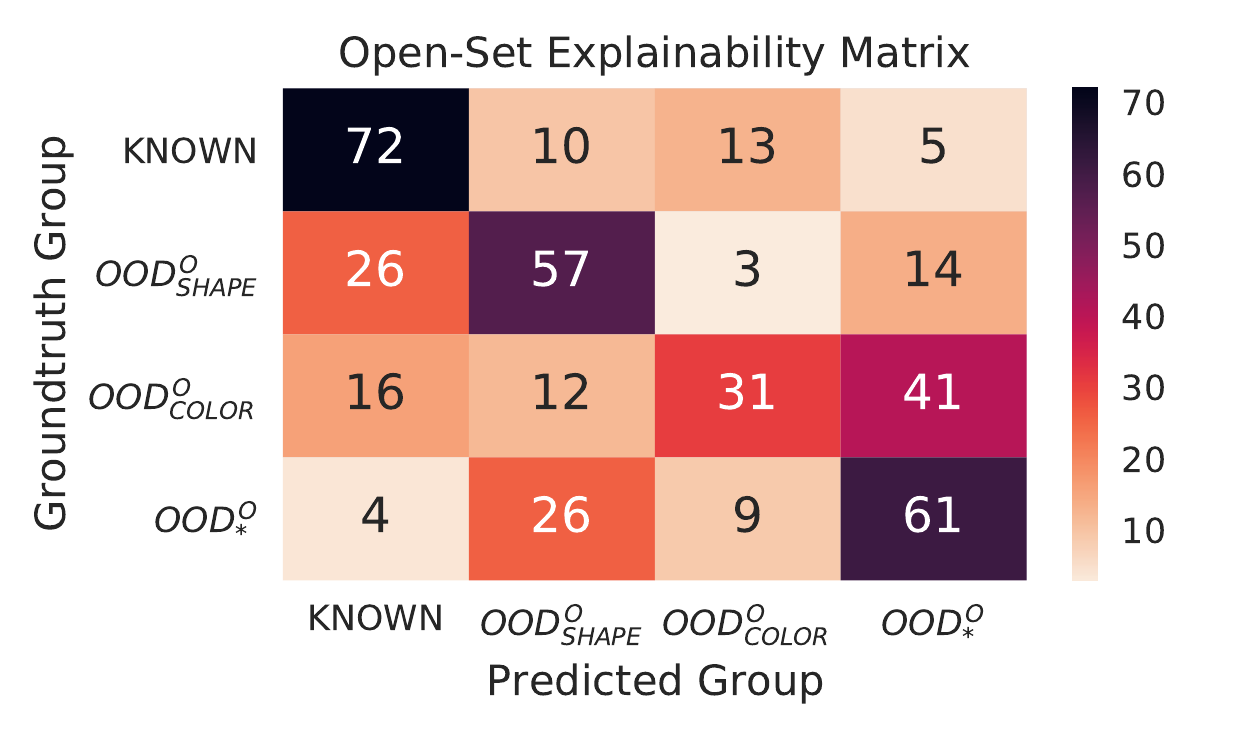}
        \label{fig:oe_dv_softmax_uc}
    }
    \subfigure[Semi-Correlated (SC)]{
        \includegraphics[width=0.3\textwidth]{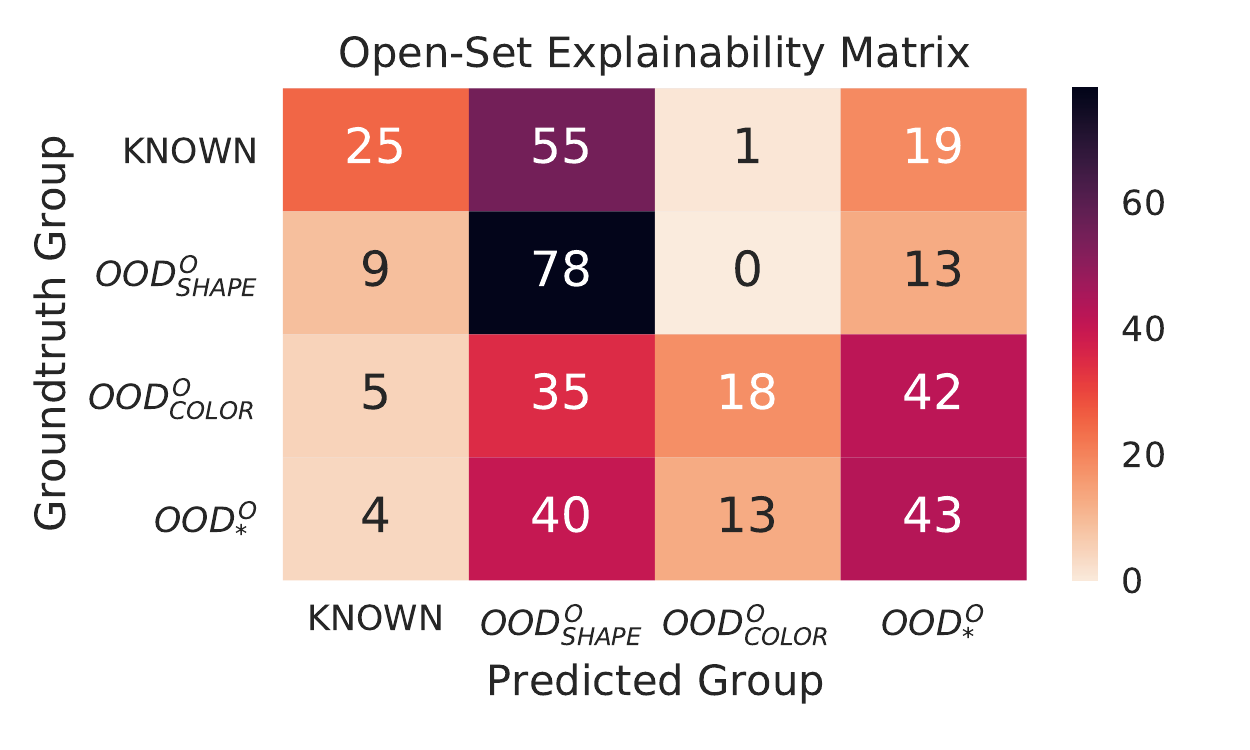}
        \label{fig:oe_dv_softmax_sc}
    }    
    \subfigure[Correlated (C)]{
        \includegraphics[width=0.3\textwidth]{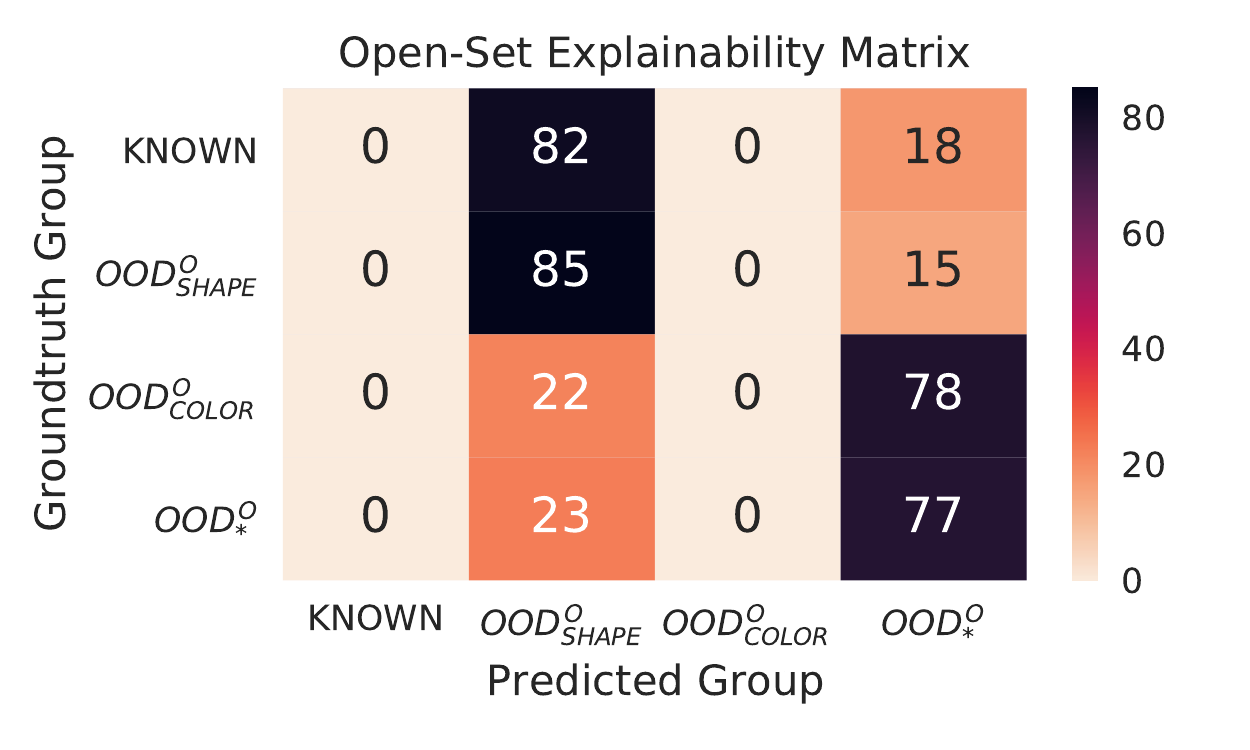}
        \label{fig:oe_dv_softmax_c}
    }
    
    \caption{Open-Set Explainability Matrix of MSP on different configurations of the Color-MNIST dataset.}
    \label{fig:oe_dv}
\end{figure*}

%% file: figures/tex/figure_oe_obc_color.tex
\begin{figure*} [!ht]
    \centering
    \subfigure[Uncorrelated (UC)]{
        \includegraphics[width=0.3\textwidth]{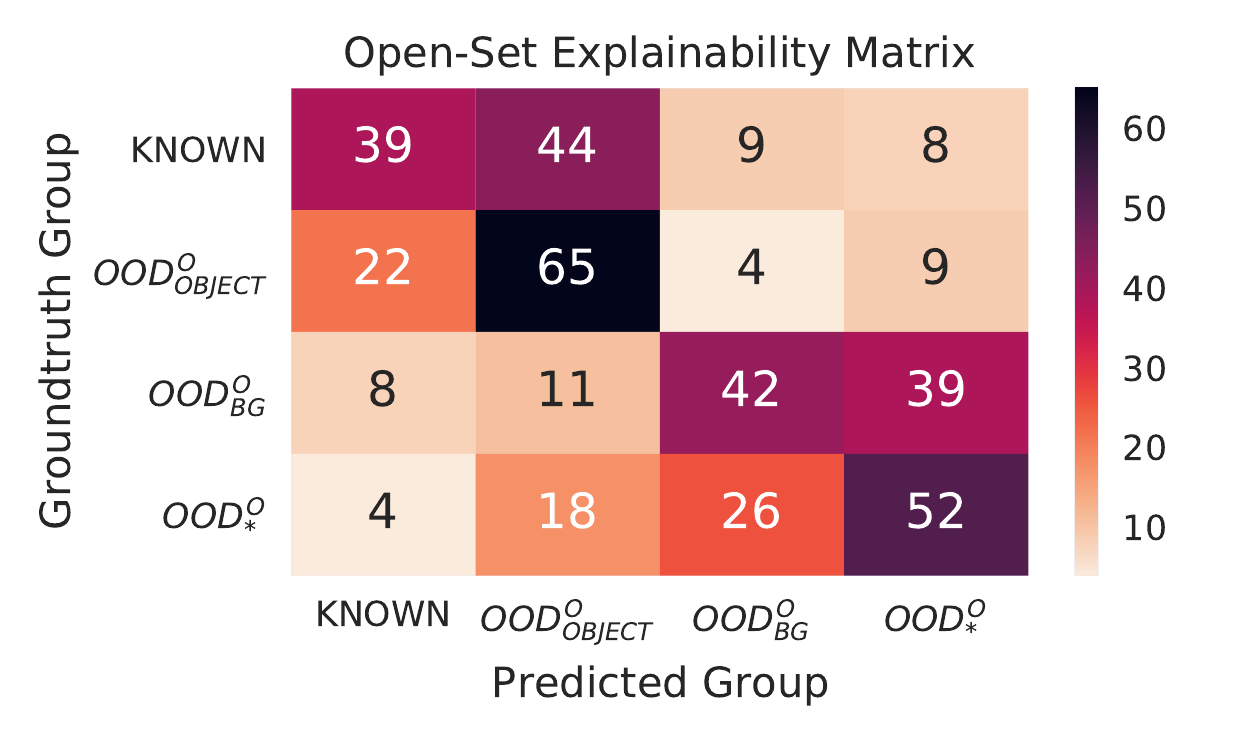}
        \label{fig:oe_obc_color_softmax_uc}
    }
    \subfigure[Semi-Correlated (SC)]{
        \includegraphics[width=0.3\textwidth]{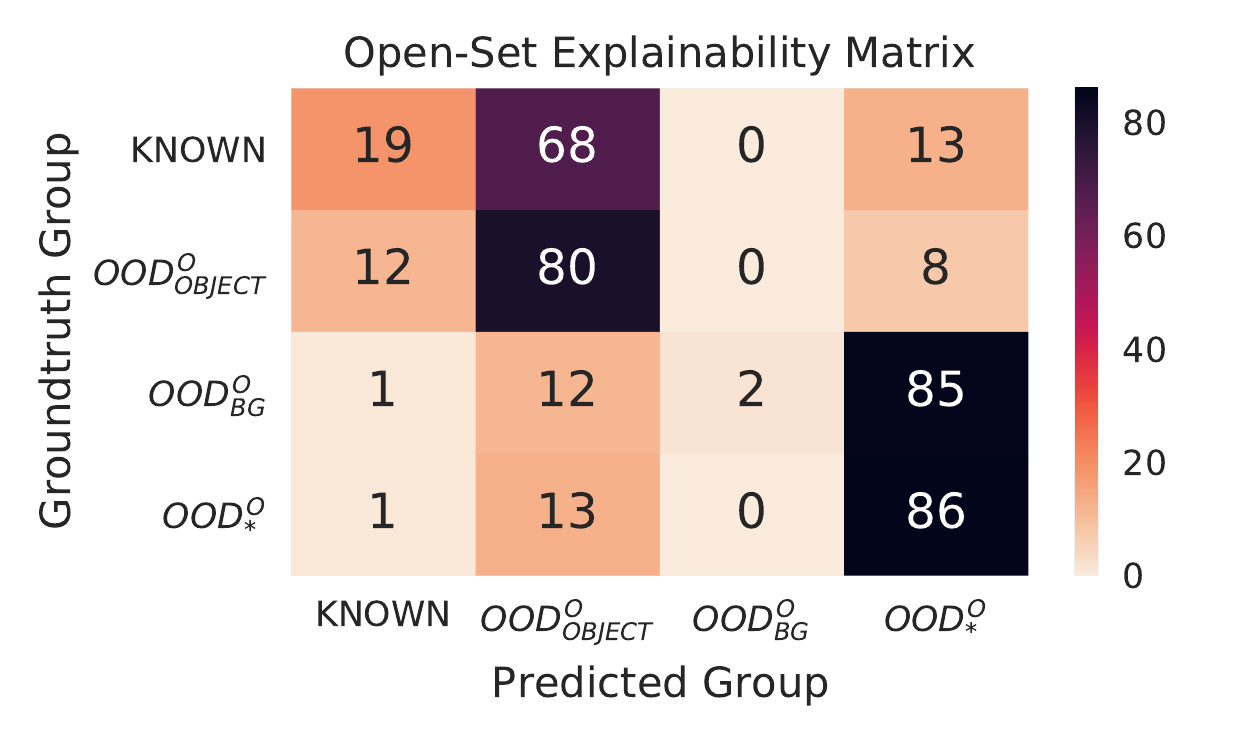}
        \label{fig:oe_obc_color_softmax_sc}
    }    
    \subfigure[Correlated (C)]{
        \includegraphics[width=0.3\textwidth]{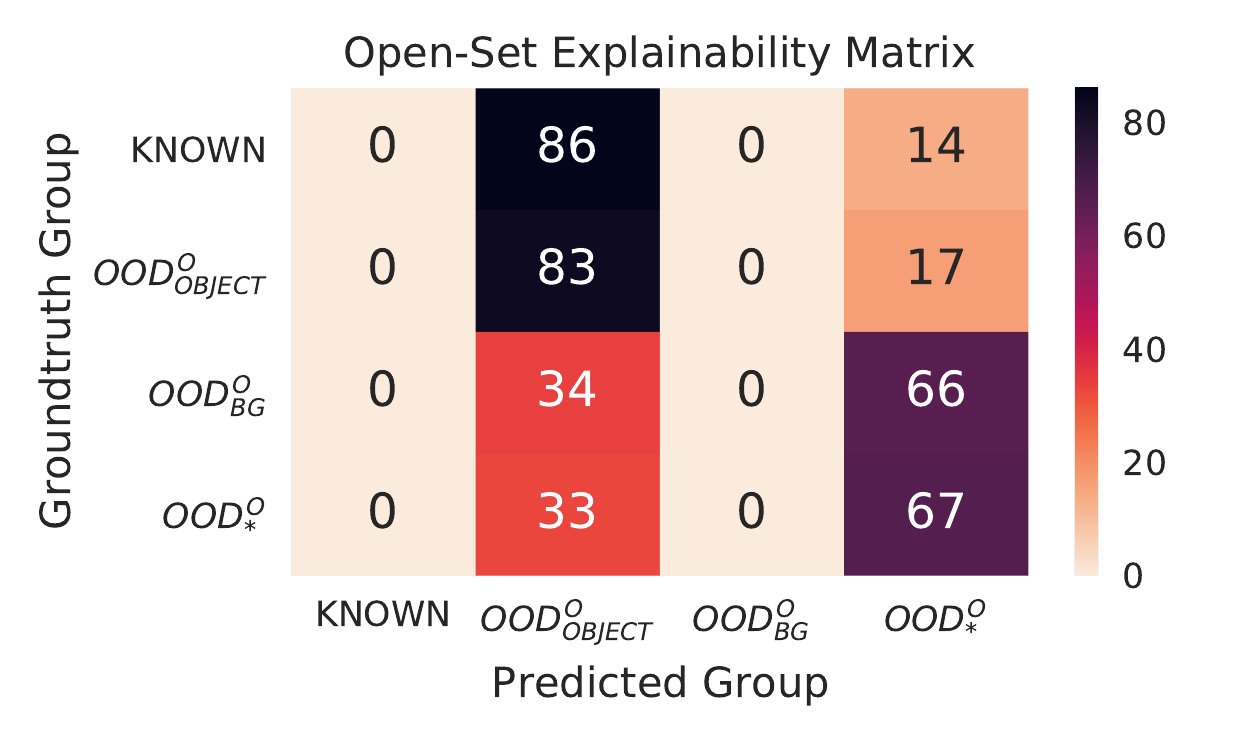}
        \label{fig:oe_obc_color_softmax_c}
    }
    
    
        \caption{Open-Set Explainability Matrix of MSP on different configurations of the Color-Object dataset.}
    \label{fig:oe_obc_color}
\end{figure*}

%% file: figures/tex/figure_evaluation_ut_zappos.tex
\begin{figure*} [t]
    \centering
    \subfigure[AUROC evaluation]{
        \begin{tabular}{l|ccc}
            \Xhline{4\arrayrulewidth}
            Approach & $\mathcal{A}_c$ & $\mathcal{A}_s$ & $\bar{\mathcal{A}}$ \\ \hline
             MSP       & $45.3$ & $67.7$ & $56.5$   \\
             OpenMax   & $45.2$ & $68.3$ & $56.8$   \\
             ARPL      & $51.0$ & $69.6$ & $60.3$   \\
             ARPL+CS   & $51.9$ & $67.5$ & $59.7$   \\
             MLS       & $44.7$ & $71.2$ & $58.0$   \\
            \Xhline{4\arrayrulewidth}
        \end{tabular}
        \label{fig:evaluation_ut_zappos_auroc_avg}
    }
    \subfigure[Confidence scores]{
        \includegraphics[width=0.3\textwidth]{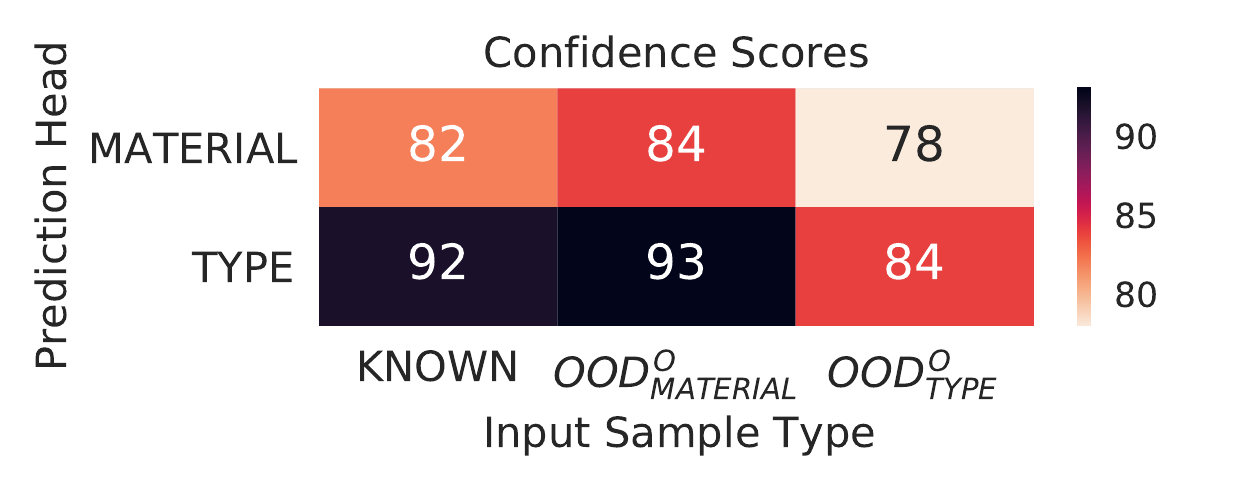}
        \label{fig:evaluation_ut_zappos_cconf_msp}
    }
    \caption{Qualitative Evaluation on UT-Zappos (a) AUROC for material ($\mathcal{A}_c$), type ($\mathcal{A}_s$) predictions and their average ($\bar{\mathcal{A}}$) (b) Cross-attribute confidence scores of MSP.}
    \label{fig:evaluation_ut_zappos}
\end{figure*}


%% file: sections/05_conclusion.tex
\section{Conclusion}
\label{sec:conclusion}

In this work, we generalize the single-task OSR to a multi-attribute OSR setup. This enables trained models to identify unknown properties of input images in an attribute-wise manner. We propose simple extensions of common OSR baselines for our task. These extended baselines are evaluated across multiple datasets. We found that these na\"ive extensions are shortcut vulnerable especially when higher degrees of attribute correlations are introduced in training sets. An investigation of how the models are affected by shortcuts is presented. We found that confidence scores predicted by the models are likely to be affected by unrelated attributes when spurious correlations are introduced. Lastly, we also evaluate the explainability of the models trained on our task. While the problem we investigated in this work remains open challenge in the domain of visual perception, we hope that it can inspire further researches in this direction.

%% file: sections/06_acknowledgements.tex



%% file: sections/07_appendix.tex
\section{Additional Quantitative Evaluation}

\subsection{Attribute-wise Open-Set Classification Rate (OSCR)}
\label{appendix:quantitative_attribute_wise_performance}

\input{tables/table_correlations_oscr_0}

\input{tables/table_correlations_oscr_1}

In section \ref{subsec:experiments_synthetic}, we have investigated the effects of spurious correlations on multi-attribute OSR performance. We mentioned that spurious correlations do not have the same effects on all attributes. Specifically, the correlations mainly degrade recognition performance of the the complex attribute $\mathcal{A}_c$ more significantly. In this section, we would like to provide quantitative evaluation in more details.

OSCR evaluation for $\mathcal{A}_c$ and $\mathcal{A}_s$ are presented in Table \ref{table:correlations_oscr_0} and \ref{table:correlations_oscr_1} respectively. It can be clearly observed from the tables that, the degradation of OSR performance due to spurious correlations on $\mathcal{A}_c$ is more severe. Considering the evaluation of Color-MNIST as an example, the OSCR of MSP for $\mathcal{A}_c$ is decreased from 76.2 in the uncorrelated case to 11.4 in the correlated case (see Table \ref{table:correlations_oscr_0}). While, for $\mathcal{A}_s$, the difference between OSCR in the uncorrelated case and the correlated case is much smaller (84.9 and 74.4 for MSP from Table \ref{table:correlations_oscr_1}). This evidence infers that, spurious correlations encourage deep networks to focus more on encoding the information of simple attribute leading to shortcut learning that degrades recognition performance of the complex attribute.

While, in Table \ref{table:correlations_oscr_0}, OSCR metrics of the complex attribute $\mathcal{A}_c$ monotonically reduce with respect to the correlations consistently, we observe that, for the simple attribute $\mathcal{A}_s$, some baselines perform better in more correlated cases. For example, according to Table \ref{table:correlations_oscr_1}, on Color-MNIST dataset, ARPL has OSCR of 87.1 in the correlated case which is higher than 82.6 in the uncorrelated case. This is not a surprise behavior as, in the correlated case, both $\mathcal{A}_c$ and $\mathcal{A}_s$ are equally predictive for target labels during training. In this regard, a model is likely to based its decision on $\mathcal{A}_s$ more than $\mathcal{A}_c$ as $\mathcal{A}_s$ is easier to learn resulting in $\mathcal{A}_s$ being a shortcut feature. Consequently, extensively emphasizing on $\mathcal{A}_s$ in the correlated cases can, for some baselines, even lead to increased performance on $\mathcal{A}_s$ as observed in Table \ref{table:correlations_oscr_1}.

\subsection{Evaluation with AUROC Metric}
\label{appendix:quantitative_auroc_avg}

\input{tables/table_correlations_auroc_avg}
\input{tables/table_correlations_auroc_0}
\input{tables/table_correlations_auroc_1}

In section \ref{subsec:experiments_synthetic}, we conduct experiments on multi-attribute OSR and observe relationship between spurious correlations and OSR performance evaluated with OSCR metric. In this section, we would like to show that using AUROC metric for evaluation provides also consistent results. According to the results in Table \ref{table:correlations_auroc_avg}, \ref{table:correlations_auroc_0} and \ref{table:correlations_auroc_1}, the following observations can still be observed:

\begin{itemize}
    \item Spurious correlations cause degradation of multi-attribute OSR performance (Table \ref{table:correlations_auroc_avg} provides similar observations as Table \ref{table:correlations_oscr_avg}). 
    \item Performance degradation on the complex attribute is more significant compared to the one on the simple attribute (\ref{table:correlations_auroc_0} and \ref{table:correlations_auroc_1} provide similar observations as Table \ref{table:correlations_oscr_0} and \ref{table:correlations_oscr_1}).
\end{itemize}


\section{Implementation Details}
\label{appendix:implementation_details}

In this section, we would like to provide additional details of our training and network hyperparameters. We would begin with the details of MSP, OpenMax and MLS baselines. These baselines have common architectures and loss terms during training. The only difference among these baselines is confident score estimation. For multi-attribute OSR, their network architectures are based on the illustration in Figure \ref{fig:architectures_multi}. For Color-MNIST, the feature extraction $g$ is LeNet followed by a linear layer producing a feature representation $z$ of 128 dimensions. Each prediction head $h_i$ is a simple multilayer perceptron (MLP) with 2 hidden layers (each has a linear layer of 128 hidden units followed by Batch normalization and ReLU activation). We use Adam optimization for training with the learning rate of $0.001$. The maximum number of training epoch is 400. For Color-Object, Scene-Object and UT-Zappos, most hyperparameters are the same as in the case of Color-MNIST except that the feature extraction $g$ is based on ResNet18 instead of LeNet. Also, learning rate of $0.0001$ is used. Our experiments are implemented using PyTorch. The code for our implementation is publicly available at \url{https://github.com/boschresearch/multiosr}.

For ARPL variations, we perform experiments based on the code and hyperparameters published by the author \cite{chen2020learning}. However, for fair comparison, we have modified its feature extraction to LeNet (for Color-MNIST) and ResNet-18 (for Color-Object, Scene-Object and UT-Zappos).

At the end of each training epoch, the loss will be computed also on the hold-out validation set. The final model evaluated on the test set will be chosen across all training epochs as the one having the minimum total loss on the validation set.





\section{Details of the Dataset Details}
\label{appendix:dataset_details}

\input{tables/table_dataset_splits_color_mnist}

\input{tables/table_dataset_splits_color_object}

\input{tables/table_dataset_splits_scene_object}

\input{tables/table_dataset_splits_ut_zappos}

In this section, we would like to provide details of the specific splits used in different dataset configurations. The details of Color-MNIST, Color-Object, Scene-Object and UT-Zappos datasets are provided in Table \ref{table:dataset_splits_color_mnist}, \ref{table:dataset_splits_color_object}, \ref{table:dataset_splits_scene_object} and \ref{table:dataset_splits_ut_zappos} respectively. For Color-Object, the attribute values of color are defined in RGB space as follows: $C_1, C_2, \ldots, C_{12}$ = (0, 100, 0), (188, 143, 143), (255, 0, 0), (255, 215, 0), (0, 255, 0), (65, 105, 225), (255, 20, 147), (135, 188, 191), (27, 145, 68), (48, 101, 53), (20, 235, 154), (187, 33, 227). 

%% file: tables/table_correlations_oscr_0.tex
\begin{table*}[b]
\caption{Average OSCR (from the complex attribute $\mathcal{A}_c$) on Color-MNIST, Color-Object and Scene-Object.}
\centering
\fontsize{9}{11}\selectfont
\begin{tabular}{l|ccc|ccc|ccc}
\Xhline{4\arrayrulewidth}
\multirow{2}{*}{Approach} & \multicolumn{3}{c|}{Color-MNIST} & \multicolumn{3}{c|}{Color-Object} & \multicolumn{3}{c}{Scene-Object} \\ 
& UC & SC & C & UC & SC & C & UC & SC & C \\ \hline
 MSP       & $76.2$       & $30.3$                 & $11.4$     & $55.3$                     & $23.0$                               & $10.4$                   & $29.9$                     & $22.0$                               & $18.9$                   \\
 OpenMax   & $76.1$       & $29.0$                 & $11.3$     & $47.4$                     & $14.2$                               & $11.1$                   & $27.6$                     & $13.1$                               & $11.1$                   \\
 ARPL      & $79.0$       & $27.3$                 & $11.2$     & $57.7$                     & $22.7$                               & $9.7$                    & $34.5$                     & $24.2$                               & $19.6$                   \\
 ARPL+CS   & $81.4$       & $28.3$                 & $11.1$     & $59.5$                     & $22.5$                               & $9.3$                    & $36.3$                     & $24.7$                               & $18.5$                   \\
 MLS       & $74.2$       & $26.2$                 & $10.9$     & $53.5$                     & $22.2$                               & $10.3$                   & $29.4$                     & $21.7$                               & $19.0$                   \\ \hline
 MSP-D     & $81.8$       & $33.3$                 & $11.4$     & $58.2$                     & $27.4$                               & $10.7$                   & $33.6$                     & $24.5$                               & $18.6$                   \\
 OpenMax-D & $81.4$       & $32.7$                 & $13.0$     & $53.4$                     & $17.0$                               & $10.4$                   & $31.5$                     & $14.5$                               & $11.1$                   \\
 ARPL-D    & $81.6$       & $31.2$                 & $11.8$     & $59.4$                     & $28.8$                               & $12.1$                   & $33.0$                     & $22.4$                               & $16.8$                   \\
 \Xhline{4\arrayrulewidth}
\end{tabular}
\label{table:correlations_oscr_0}
\end{table*}

%% file: tables/table_correlations_oscr_1.tex
\begin{table*}[t]
\caption{Average OSCR (from the simple attribute $\mathcal{A}_s$) on Color-MNIST, Color-Object and Scene-Object.}
\centering
\fontsize{9}{11}\selectfont
\begin{tabular}{l|ccc|ccc|ccc}
\Xhline{4\arrayrulewidth}
\multirow{2}{*}{Approach} & \multicolumn{3}{c|}{Color-MNIST} & \multicolumn{3}{c|}{Color-Object} & \multicolumn{3}{c}{Scene-Object} \\ 
& UC & SC & C & UC & SC & C & UC & SC & C \\ \hline
 MSP       & $84.9$       & $81.8$                 & $74.4$     & $97.1$                     & $93.8$                               & $95.7$                   & $49.5$                     & $42.3$                               & $39.6$                   \\
 OpenMax   & $82.2$       & $76.7$                 & $79.1$     & $93.0$                     & $77.3$                               & $38.9$                   & $43.5$                     & $23.8$                               & $15.0$                   \\
 ARPL      & $82.6$       & $80.8$                 & $87.1$     & $97.0$                     & $86.2$                               & $62.4$                   & $50.4$                     & $44.7$                               & $40.3$                   \\
 ARPL+CS   & $75.9$       & $77.1$                 & $88.7$     & $95.6$                     & $85.2$                               & $67.7$                   & $53.0$                     & $44.8$                               & $38.1$                   \\
 MLS       & $78.8$       & $62.9$                 & $61.0$     & $96.4$                     & $91.7$                               & $93.0$                   & $51.4$                     & $42.8$                               & $39.9$                   \\ \hline
 MSP-D     & $78.3$       & $83.3$                 & $81.5$     & $91.8$                     & $96.0$                               & $94.2$                   & $52.2$                     & $46.4$                               & $39.0$                   \\
 OpenMax-D & $72.1$       & $65.6$                 & $74.6$     & $97.0$                     & $80.0$                               & $34.8$                   & $45.8$                     & $18.7$                               & $13.3$                   \\
 ARPL-D    & $85.7$       & $89.4$                 & $79.8$     & $98.9$                     & $89.6$                               & $87.5$                   & $55.0$                     & $43.8$                               & $35.8$                   \\
 \Xhline{4\arrayrulewidth}
\end{tabular}
\label{table:correlations_oscr_1}
\end{table*}

%% file: tables/table_correlations_auroc_avg.tex
\begin{table*}[ht]
\caption{Average AUROC (from all attributes) on Color-MNIST, Color-Object and Scene-Object.}
\centering
\fontsize{9}{11}\selectfont
\begin{tabular}{l|ccc|ccc|ccc}
\Xhline{4\arrayrulewidth}
\multirow{2}{*}{Approach} & \multicolumn{3}{c|}{Color-MNIST} & \multicolumn{3}{c|}{Color-Object} & \multicolumn{3}{c}{Scene-Object} \\ 
& UC & SC & C & UC & SC & C & UC & SC & C \\ \hline
 MSP       & $81.9$       & $69.6$                 & $61.5$     & $82.4$                     & $73.9$                               & $74.0$                   & $58.3$                     & $57.3$                               & $57.2$                   \\
 OpenMax   & $80.8$       & $66.8$                 & $64.0$     & $78.8$                     & $67.8$                               & $55.4$                   & $57.2$                     & $52.2$                               & $52.0$                   \\
 ARPL      & $83.3$       & $69.8$                 & $67.7$     & $82.5$                     & $69.1$                               & $57.4$                   & $59.2$                     & $57.7$                               & $58.2$                   \\
 ARPL+CS   & $79.8$       & $67.1$                 & $68.4$     & $82.3$                     & $68.9$                               & $59.4$                   & $61.5$                     & $57.8$                               & $57.4$                   \\
 MLS       & $78.2$       & $58.2$                 & $54.7$     & $81.5$                     & $72.6$                               & $72.6$                   & $59.4$                     & $58.1$                               & $57.7$                   \\ \hline
 MSP-D     & $80.8$       & $71.4$                 & $65.0$     & $80.3$                     & $74.9$                               & $72.9$                   & $60.1$                     & $59.1$                               & $57.2$                   \\
 OpenMax-D & $77.4$       & $61.5$                 & $62.1$     & $80.1$                     & $65.1$                               & $55.5$                   & $57.9$                     & $51.6$                               & $50.5$                   \\
 ARPL-D    & $85.3$       & $73.3$                 & $64.1$     & $83.9$                     & $72.1$                               & $70.5$                   & $61.2$                     & $57.7$                               & $57.1$                   \\
\Xhline{4\arrayrulewidth}
\end{tabular}
\label{table:correlations_auroc_avg}
\end{table*}

%% file: tables/table_correlations_auroc_0.tex
\begin{table*}[ht]
\caption{Average AUROC (from the complex attribute $\mathcal{A}_c$) on Color-MNIST, Color-Object and Scene-Object.}
\centering
\fontsize{9}{11}\selectfont
\begin{tabular}{c|ccc|ccc|ccc}
\Xhline{4\arrayrulewidth}
\multirow{2}{*}{Approach} & \multicolumn{3}{c|}{Color-MNIST} & \multicolumn{3}{c|}{Colored-Objects} & \multicolumn{3}{c}{Scene-Objects} \\ 
& UC & SC & C & UC & SC & C & UC & SC & C \\ \hline
 MSP       & $78.9$       & $57.4$                 & $48.6$     & $67.4$                     & $53.9$                               & $52.2$                   & $54.4$                     & $55.3$                               & $53.2$                   \\
 OpenMax   & $79.4$       & $56.9$                 & $48.3$     & $63.1$                     & $52.1$                               & $50.0$                   & $53.6$                     & $51.1$                               & $51.5$                   \\
 ARPL      & $84.1$       & $58.9$                 & $48.3$     & $67.9$                     & $51.9$                               & $52.3$                   & $56.8$                     & $54.5$                               & $54.7$                   \\
 ARPL+CS   & $83.7$       & $57.1$                 & $48.0$     & $68.9$                     & $52.5$                               & $51.1$                   & $58.8$                     & $54.4$                               & $53.2$                   \\
 MLS       & $77.5$       & $53.6$                 & $48.3$     & $66.4$                     & $53.3$                               & $52.0$                   & $53.7$                     & $55.4$                               & $53.2$                   \\ \hline
 MSP-D     & $83.3$       & $59.6$                 & $48.5$     & $68.7$                     & $53.8$                               & $51.5$                   & $56.8$                     & $56.1$                               & $54.1$                   \\
 OpenMax-D & $82.7$       & $57.3$                 & $49.2$     & $64.5$                     & $49.6$                               & $50.6$                   & $55.7$                     & $51.3$                               & $50.8$                   \\
  ARPL-D    & $84.9$       & $57.2$                 & $48.4$     & $68.9$                     & $54.4$                               & $52.5$                   & $56.1$                     & $54.0$                               & $54.0$                   \\
 \Xhline{4\arrayrulewidth}
\end{tabular}
\label{table:correlations_auroc_0}
\end{table*}

%% file: tables/table_correlations_auroc_1.tex
\begin{table*}[ht]
\caption{Average AUROC (from the simple attribute $\mathcal{A}_s$) on Color-MNIST, Color-Object and Scene-Object.}
\centering
\fontsize{9}{11}\selectfont
\begin{tabular}{c|ccc|ccc|ccc}
\Xhline{4\arrayrulewidth}
\multirow{2}{*}{Approach} & \multicolumn{3}{c|}{Color-MNIST} & \multicolumn{3}{c|}{Colored-Objects} & \multicolumn{3}{c}{Scene-Objects} \\ 
& UC & SC & C & UC & SC & C & UC & SC & C \\ \hline
 MSP       & $85.0$       & $81.8$                 & $74.4$     & $97.3$                     & $93.9$                               & $95.7$                   & $62.2$                     & $59.3$                               & $61.2$                   \\
 OpenMax   & $82.2$       & $76.8$                 & $79.8$     & $94.4$                     & $83.6$                               & $60.8$                   & $60.9$                     & $53.3$                               & $52.4$                   \\
 ARPL      & $82.6$       & $80.8$                 & $87.1$     & $97.2$                     & $86.4$                               & $62.5$                   & $61.5$                     & $60.8$                               & $61.6$                   \\
 ARPL+CS   & $75.9$       & $77.1$                 & $88.7$     & $95.7$                     & $85.3$                               & $67.8$                   & $64.2$                     & $61.2$                               & $61.6$                   \\
 MLS       & $78.8$       & $62.9$                 & $61.0$     & $96.6$                     & $91.9$                               & $93.1$                   & $65.2$                     & $60.7$                               & $62.3$                   \\ \hline
 MSP-D     & $78.2$       & $83.3$                 & $81.6$     & $91.9$                     & $96.1$                               & $94.3$                   & $63.3$                     & $62.0$                               & $60.3$                   \\
 OpenMax-D & $72.1$       & $65.7$                 & $74.9$     & $95.8$                     & $80.6$                               & $60.5$                   & $60.1$                     & $51.8$                               & $50.2$                   \\
 ARPL-D    & $85.7$       & $89.4$                 & $79.8$     & $98.9$                     & $89.8$                               & $88.6$                   & $66.3$                     & $61.4$                               & $60.2$                   \\
 \Xhline{4\arrayrulewidth}
\end{tabular}
\label{table:correlations_auroc_1}
\end{table*}

%% file: tables/table_dataset_splits_color_mnist.tex
\begin{table*}[t]
\caption{Details of Color-MNIST dataset splits.}
\centering
\fontsize{9}{11}\selectfont
\begin{tabular}{l|>{\centering\arraybackslash}m{0.6\linewidth}}
\Xhline{4\arrayrulewidth}
Property & Details \\ \hline
Complex/Simple Attributes & Digit / Color \\ \hline
Known Attribute Values & \{0, 1, 2, 3, 4\} / \newline \{red, yellow, green, cyan, blue\} \\ \hline
Unknown Attribute Values & \{5, 6, 7, 8, 9\} / \newline \{magenta, orange, violet, azure, rose\} \\ \hline
\makecell[l]{Attribute Combinations \\ (Correlated Training)} & \{(0, red), (1, yellow), (2, green), (3, cyan), (4, blue)\} \\ \hline
\makecell[l]{Attribute Combinations \\ (Semi-Correlated Training)} & \{(0, red), (0, yellow), (1, yellow), (1, green), (2, green), (2, cyan), (3, cyan), (3, blue), (4, blue), (4, yellow)\} \\ \hline
 \Xhline{4\arrayrulewidth}
\end{tabular}
\label{table:dataset_splits_color_mnist}
\end{table*}

%% file: tables/table_dataset_splits_color_object.tex
\begin{table*}[t]
\caption{Details of Color-Object dataset splits.}
\centering
\fontsize{9}{11}\selectfont
\begin{tabular}{l|>{\centering\arraybackslash}m{0.6\linewidth}}
\Xhline{4\arrayrulewidth}
Property & Details \\ \hline
Complex/Simple Attributes & Object Type / Color\\ \hline
Known Attribute Values & \{boat, airplane, truck, dog, zebra, horse\} / \newline \{$C_1$,$C_2$, $C_3$, $C_4$, $C_5$, $C_6$\} \\ \hline
Unknown Attribute Values & \{bird, motorcycle, elephant, bear, bed, giraffe\} / \newline \{$C_7$, $C_8$, $C_9$, $C_{10}$, $C_{11}$, $C_{12}$\} \\ \hline
\makecell[l]{Attribute Combinations \\ (Correlated Training)} & \{(boat, $C_1$), (airplane, $C_2$), (truck, $C_3$), (dog, $C_4$), (zebra, $C_5$), (horse, $C_6$)\} \\ \hline
\makecell[l]{Attribute Combinations \\ (Semi-Correlated Training)} & \{(boat, $C_1$), \{(boat, $C_2$), (airplane, $C_2$), (airplane, $C_3$), (truck, $C_3$), (truck, $C_4$), (dog, $C_4$), (dog, $C_5$), (zebra, $C_5$), (zebra, $C_6$), (horse, $C_6$), (horse, $C_1$)\} \\ \hline
 \Xhline{4\arrayrulewidth}
\end{tabular}
\label{table:dataset_splits_color_object}
\end{table*}

%% file: tables/table_dataset_splits_scene_object.tex
\begin{table*}[t]
\caption{Details of Scene-Object dataset splits.}
\centering
\fontsize{9}{11}\selectfont
\begin{tabular}{l|>{\centering\arraybackslash}m{0.6\linewidth}}
\Xhline{4\arrayrulewidth}
Property & Details \\ \hline
Complex/Simple Attributes & Object Type / Scene Type\\ \hline
Known Attribute Values & \{boat, airplane, truck, dog, zebra, horse\} / \newline \{beach, canyon, building, stair, desert, crevasse\} \\ \hline
Unknown Attribute Values & \{bird, motorcycle, elephant, bear, bed, giraffe\} / \newline \{ball\_pit, oast\_house, kasbah, lighthouse, pagoda, rock\_arch\} \\ \hline
\makecell[l]{Attribute Combinations \\ (Correlated Training)} & \{(boat, beach), (airplane, canyon), (truck, building), (dog, stair), (zebra, desert), (horse, crevasse)\} \\ \hline
\makecell[l]{Attribute Combinations \\ (Semi-Correlated Training)} & \{(boat, beach), \{(boat, canyon), (airplane, canyon), (airplane, building), (truck, building), (truck, stair), (dog, stair), (dog, desert), (zebra, desert), (zebra, crevasse), (horse, crevasse), (horse, beach)\} \\ \hline
 \Xhline{4\arrayrulewidth}
\end{tabular}
\label{table:dataset_splits_scene_object}
\end{table*}

%% file: tables/table_dataset_splits_ut_zappos.tex
\begin{table*}[t]
\caption{Details of UT-Zappos dataset splits.}
\centering
\fontsize{9}{11}\selectfont
\begin{tabular}{l|>{\centering\arraybackslash}m{0.6\linewidth}}
\Xhline{4\arrayrulewidth}
Property & Details \\ \hline
Complex/Simple Attributes & Shoe Material / Shoe Type \\ \hline
Known Attribute Values & \{Faux.Leather, Full.grain.leather, Leather, Rubber, Suede\} / \newline \{Boots.Knee.High, Boots.Mid-Calf, Shoes.Flats, Shoes.Heels, Shoes.Loafers\} \\ \hline
Unknown Attribute Values & \{Canvas, Nubuck, Patent.Leather, Satin, Synthetic\} / \newline \{Boots.Ankle, Sandals, Shoes.Oxfords, Shoes.Sneakers.and.Athletic.Shoes\} \\ \hline
\makecell[l]{Attribute Combinations \\ (Training)} & \{(Faux.Leather, Boots.Knee.High), (Faux.Leather, Boots.Mid-Calf), (Faux.Leather, Shoes.Flats), (Full.grain.leather, Boots.Mid-Calf), (Full.grain.leather, Shoes.Loafers), (Leather, Shoes.Flats), (Leather, Shoes.Heels), (Leather, Shoes.Loafers), (Rubber, Boots.Knee.High), (Rubber, Boots.Mid-Calf), (Suede, Boots.Knee.High), (Suede, Shoes.Flats), (Suede, Shoes.Heels)\} \\ \hline
 \Xhline{4\arrayrulewidth}
\end{tabular}
\label{table:dataset_splits_ut_zappos}
\end{table*}

%% file: main.bbl
\begin{thebibliography}{10}
\providecommand{\url}[1]{\texttt{#1}}
\providecommand{\urlprefix}{URL }
\providecommand{\doi}[1]{https://doi.org/#1}

\bibitem{ahmed2020systematic}
Ahmed, F., Bengio, Y., van Seijen, H., Courville, A.: Systematic generalisation
  with group invariant predictions. In: International Conference on Learning
  Representations (2020)

\bibitem{atzmon2020causal}
Atzmon, Y., Kreuk, F., Shalit, U., Chechik, G.: A causal view of compositional
  zero-shot recognition. In: Larochelle, H., Ranzato, M., Hadsell, R., Balcan,
  M.F., Lin, H. (eds.) Advances in Neural Information Processing Systems.
  vol.~33, pp. 1462--1473. Curran Associates, Inc. (2020)

\bibitem{bendale2016towards}
Bendale, A., Boult, T.E.: Towards open set deep networks. In: Proceedings of
  the IEEE conference on computer vision and pattern recognition. pp.
  1563--1572 (2016)

\bibitem{chen2021adversarial}
Chen, G., Peng, P., Wang, X., Tian, Y.: Adversarial reciprocal points learning
  for open set recognition. arXiv preprint arXiv:2103.00953  (2021)

\bibitem{chen2020learning}
Chen, G., Qiao, L., Shi, Y., Peng, P., Li, J., Huang, T., Pu, S., Tian, Y.:
  Learning open set network with discriminative reciprocal points. In: European
  Conference on Computer Vision. pp. 507--522. Springer (2020)

\bibitem{du2022class}
Du, S., Hong, C., Chen, Y., Cao, Z., Zhang, Z.: Class-attribute inconsistency
  learning for novelty detection. Pattern Recognition  \textbf{126},  108582
  (2022)

\bibitem{eulig2021diagvib}
Eulig, E., Saranrittichai, P., Mummadi, C.K., Rambach, K., Beluch, W., Shi, X.,
  Fischer, V.: Diagvib-6: A diagnostic benchmark suite for vision models in the
  presence of shortcut and generalization opportunities. In: Proceedings of the
  IEEE/CVF International Conference on Computer Vision. pp. 10655--10664 (2021)

\bibitem{ge2017generative}
Ge, Z., Demyanov, S., Chen, Z., Garnavi, R.: Generative openmax for multi-class
  open set classification. arXiv preprint arXiv:1707.07418  (2017)

\bibitem{geirhos2020shortcut}
Geirhos, R., Jacobsen, J.H., Michaelis, C., Zemel, R., Brendel, W., Bethge, M.,
  Wichmann, F.A.: Shortcut learning in deep neural networks. Nature Machine
  Intelligence  \textbf{2}(11),  665--673 (2020)

\bibitem{geirhos2018imagenet}
Geirhos, R., Rubisch, P., Michaelis, C., Bethge, M., Wichmann, F.A., Brendel,
  W.: Imagenet-trained {CNN}s are biased towards texture; increasing shape bias
  improves accuracy and robustness. In: ICLR (2018)

\bibitem{geng2020recent}
Geng, C., Huang, S.j., Chen, S.: Recent advances in open set recognition: A
  survey. IEEE transactions on pattern analysis and machine intelligence
  \textbf{43}(10),  3614--3631 (2020)

\bibitem{gillert2021towards}
Gillert, A., von Lukas, U.F.: Towards combined open set recognition and
  out-of-distribution detection for fine-grained classification. In: VISIGRAPP
  (5: VISAPP). pp. 225--233 (2021)

\bibitem{guo2021conditional}
Guo, Y., Camporese, G., Yang, W., Sperduti, A., Ballan, L.: Conditional
  variational capsule network for open set recognition. arXiv preprint
  arXiv:2104.09159  (2021)

\bibitem{hermann2020shapes}
Hermann, K., Lampinen, A.: What shapes feature representations? exploring
  datasets, architectures, and training. In: NeurIPS. vol.~33, pp. 9995--10006.
  Curran Associates, Inc. (2020)

\bibitem{lin2014microsoft}
Lin, T.Y., Maire, M., Belongie, S., Hays, J., Perona, P., Ramanan, D.,
  Doll{\'a}r, P., Zitnick, C.L.: Microsoft coco: Common objects in context. In:
  European conference on computer vision. pp. 740--755. Springer (2014)

\bibitem{mancini2021open}
Mancini, M., Naeem, M.F., Xian, Y., Akata, Z.: Open world compositional
  zero-shot learning. In: Proceedings of the IEEE/CVF Conference on Computer
  Vision and Pattern Recognition. pp. 5222--5230 (2021)

\bibitem{misra2017red}
Misra, I., Gupta, A., Hebert, M.: From red wine to red tomato: Composition with
  context. In: Proceedings of the IEEE Conference on Computer Vision and
  Pattern Recognition. pp. 1792--1801 (2017)

\bibitem{naeem2021learning}
Naeem, M.F., Xian, Y., Tombari, F., Akata, Z.: Learning graph embeddings for
  compositional zero-shot learning. In: Proceedings of the IEEE/CVF Conference
  on Computer Vision and Pattern Recognition. pp. 953--962 (2021)

\bibitem{nagarajan2018attributes}
Nagarajan, T., Grauman, K.: Attributes as operators: factorizing unseen
  attribute-object compositions. In: Proceedings of the European Conference on
  Computer Vision (ECCV). pp. 169--185 (2018)

\bibitem{neal2018open}
Neal, L., Olson, M., Fern, X., Wong, W.K., Li, F.: Open set learning with
  counterfactual images. In: Proceedings of the European Conference on Computer
  Vision (ECCV). pp. 613--628 (2018)

\bibitem{oza2019c2ae}
Oza, P., Patel, V.M.: C2ae: Class conditioned auto-encoder for open-set
  recognition. In: Proceedings of the IEEE/CVF Conference on Computer Vision
  and Pattern Recognition. pp. 2307--2316 (2019)

\bibitem{purushwalkam2019task}
Purushwalkam, S., Nickel, M., Gupta, A., Ranzato, M.: Task-driven modular
  networks for zero-shot compositional learning. In: Proceedings of the
  IEEE/CVF International Conference on Computer Vision. pp. 3593--3602 (2019)

\bibitem{walter2016towards}
Scheirer, W.J., de~Rezende~Rocha, A., Sapkota, A., Boult, T.E.: Toward open set
  recognition. IEEE Transactions on Pattern Analysis and Machine Intelligence
  \textbf{35}(7),  1757--1772 (2013). \doi{10.1109/TPAMI.2012.256}

\bibitem{shu2020p}
Shu, Y., Shi, Y., Wang, Y., Huang, T., Tian, Y.: p-odn: prototype-based open
  deep network for open set recognition. Scientific reports  \textbf{10}(1),
  1--13 (2020)

\bibitem{sun2020conditional}
Sun, X., Yang, Z., Zhang, C., Ling, K.V., Peng, G.: Conditional gaussian
  distribution learning for open set recognition. In: Proceedings of the
  IEEE/CVF Conference on Computer Vision and Pattern Recognition. pp.
  13480--13489 (2020)

\bibitem{vaze2021open}
Vaze, S., Han, K., Vedaldi, A., Zisserman, A.: Open-set recognition: A good
  closed-set classifier is all you need. In: International Conference on
  Learning Representations (2021)

\bibitem{yoshihashi2019classification}
Yoshihashi, R., Shao, W., Kawakami, R., You, S., Iida, M., Naemura, T.:
  Classification-reconstruction learning for open-set recognition. In:
  Proceedings of the IEEE Conference on Computer Vision and Pattern
  Recognition. pp. 4016--4025 (2019)

\bibitem{yu2014fine}
Yu, A., Grauman, K.: Fine-grained visual comparisons with local learning. In:
  Proceedings of the IEEE Conference on Computer Vision and Pattern
  Recognition. pp. 192--199 (2014)

\bibitem{yu2017semantic}
Yu, A., Grauman, K.: Semantic jitter: Dense supervision for visual comparisons
  via synthetic images. In: Proceedings of the IEEE International Conference on
  Computer Vision. pp. 5570--5579 (2017)

\bibitem{zhang2020hybrid}
Zhang, H., Li, A., Guo, J., Guo, Y.: Hybrid models for open set recognition.
  In: European Conference on Computer Vision. pp. 102--117. Springer (2020)

\bibitem{zhou2017places}
Zhou, B., Lapedriza, A., Khosla, A., Oliva, A., Torralba, A.: Places: A 10
  million image database for scene recognition. IEEE transactions on pattern
  analysis and machine intelligence  \textbf{40}(6),  1452--1464 (2017)

\bibitem{zhou2021learning}
Zhou, D.W., Ye, H.J., Zhan, D.C.: Learning placeholders for open-set
  recognition. In: Proceedings of the IEEE/CVF Conference on Computer Vision
  and Pattern Recognition. pp. 4401--4410 (2021)

\end{thebibliography}
